\documentclass[10pt,twocolumn,letterpaper]{article}

\usepackage{cvpr}              

\usepackage{graphicx}
\usepackage{amsmath}
\usepackage{amssymb}
\usepackage{booktabs}
\usepackage{algorithm}
\usepackage{algorithmic}
\usepackage{times}
\usepackage{epsfig}
\usepackage{graphicx}
\usepackage{amssymb}
\usepackage{float}
\usepackage{color}

\usepackage{bm}
\usepackage{textcomp}
\usepackage{multirow}
\usepackage{wrapfig}
\usepackage{caption}
\usepackage{algorithm}
\usepackage{algorithmic}
\usepackage{amssymb}
\usepackage{booktabs}
\usepackage{textcomp}
\usepackage{multirow}
\usepackage{wrapfig}
\usepackage{caption}
\usepackage{comment}
\usepackage{amsthm}
\usepackage[utf8]{inputenc} 
\usepackage[T1]{fontenc}    
\usepackage{url}            
\usepackage{booktabs}       
\usepackage{amsfonts}       
\usepackage{nicefrac}       
\usepackage{microtype}      
\usepackage{times}
\usepackage{epsfig}
\usepackage{ulem}
\usepackage{graphicx}
\usepackage{amssymb}
\usepackage{amsfonts}
\usepackage{multirow}
\usepackage{algorithm}
\usepackage{algorithmic}
\usepackage{times}
\usepackage{enumitem}
\usepackage{multicol}
\usepackage{graphicx}
\usepackage{amssymb}
\usepackage{multicol}
\usepackage{bm}
\usepackage{verbatim}
\usepackage{algorithm}
\usepackage{algorithmic}
\usepackage{amssymb}
\usepackage{booktabs}

\usepackage{comment}
\usepackage{amsthm}
\usepackage{comment}
\usepackage{mathrsfs}
\usepackage{wrapfig}
\usepackage{color}

\usepackage[american]{babel}

\makeatletter

\newcommand{\qileft}{[\kern-0.15em[}
\newcommand{\qiLeft}{\left[\kern-0.4em\left[}
\newcommand{\qiright}{]\kern-0.15em]}
\newcommand{\qiRight}{\right]\kern-0.4em\right]}

\renewcommand{\Roman}[1]{\uppercase\expandafter{\romannumeral#1}}
\newcommand{\red}[1]{{\color{red}{#1}}}
\newcommand{\blue}[1]{{\color{blue}{#1}}}

\newcommand{\A}{{\mathcal{A}}}

\newcommand{\W}{\mathcal{W}}
\newcommand{\Loss}{\mathcal{L}}

\usepackage{pifont}

\newcommand{\cmark}{\ding{51}\xspace}%
\newcommand{\xmark}{\ding{55}\xspace}%
\usepackage{amssymb}

\usepackage[pagebackref,breaklinks,colorlinks]{hyperref}

\usepackage[capitalize]{cleveref}
\crefname{section}{Sec.}{Secs.}
\Crefname{section}{Section}{Sections}
\Crefname{table}{Table}{Tables}
\crefname{table}{Tab.}{Tabs.}

\usepackage{listings} 

\usepackage{color}
\definecolor{deepblue}{rgb}{0,0,0.5}
\definecolor{deepred}{rgb}{0.6,0,0}
\definecolor{deepgreen}{rgb}{0,0.5,0}
\definecolor{codebrown}{rgb}{0.8,0.44,0.2}
\definecolor{codegray}{rgb}{0.5,0.5,0.5}
\definecolor{codepurple}{rgb}{0.58,0,0.82}
\definecolor{backcolour}{rgb}{0.95,0.95,0.92}

\def\cvprPaperID{98} 
\def\confName{CVPR}
\def\confYear{2023}

\def\ourmethod{DisWOT}

\begin{document}
\title{DisWOT: Student Architecture Search for Distillation WithOut Training}

\author{Peijie Dong$^{1 \dag}$\quad Lujun Li$^{2 \dag}$\thanks{Corresponding author, $\dag$ equal contribution, PD conducted main experiments, LL proposed ideas and led the project \& writing. }\quad Zimian Wei$^{1 \dag}$\\
$^1$ National University of Defense Technology, $^2$ Chinese Academy of Sciences\\
{\tt\small $^{1}$\{dongpeijie,  weizimian16\}@nudt.edu.cn, $^{2}$lilujunai@gmail.com}
}

\maketitle

\begin{abstract}
Knowledge distillation (KD) is an effective training strategy to improve the lightweight student models under the guidance of cumbersome teachers. However, the large architecture difference across the teacher-student pairs limits the distillation gains. In contrast to previous adaptive distillation methods to reduce the teacher-student gap,  we explore a novel training-free framework to search for the best student architectures for a given teacher. Our work first empirically show that the optimal model under vanilla training cannot be the winner in distillation. Secondly, we find that the similarity of feature semantics and sample relations between random-initialized teacher-student networks have good correlations with final distillation performances. Thus, we efficiently measure similarity matrixs conditioned on the semantic activation maps to select the optimal student via an evolutionary algorithm without any training. In this way, our student architecture search for Distillation WithOut Training (DisWOT) significantly improves the performance of the model in the distillation stage with at least 180$\times$ training acceleration. Additionally,  we extend similarity metrics in DisWOT as new distillers and KD-based zero-proxies. Our experiments on CIFAR, ImageNet and NAS-Bench-201 demonstrate that our technique achieves state-of-the-art results on different search spaces. Our project and code are available at https://lilujunai.github.io/DisWOT-CVPR2023/.

\end{abstract}

\section{Introduction}
\label{sec:intro}
\begin{figure}
\setlength{\abovecaptionskip}{0.cm}
\setlength{\belowcaptionskip}{-0.cm}
\centering
  \begin{minipage}[t]{0.49\linewidth}
    \centering
\includegraphics[width=1\linewidth]{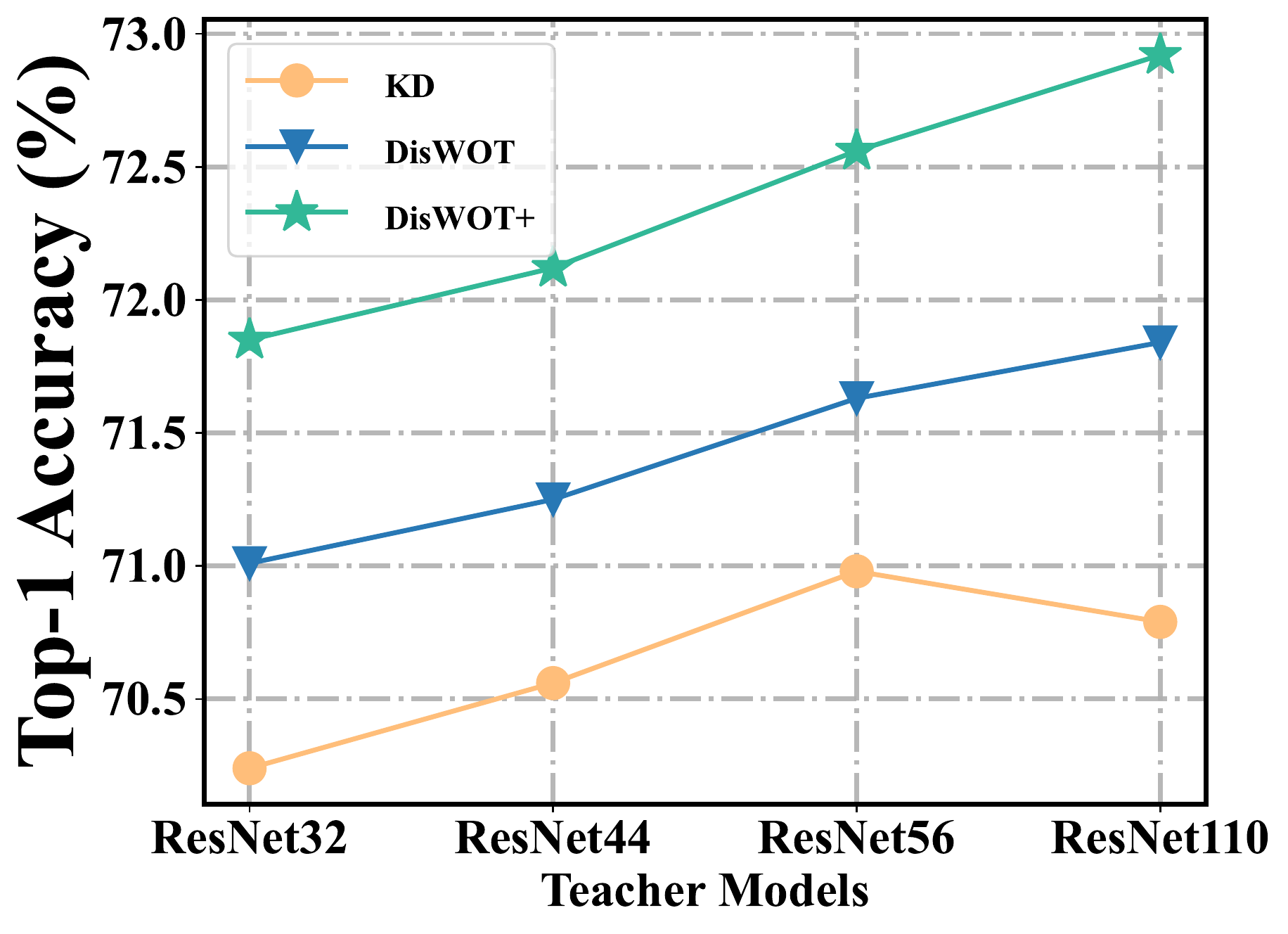}
  \end{minipage}%
  \begin{minipage}[t]{0.51\linewidth}
    \centering
\centering
\includegraphics[width=1\linewidth]{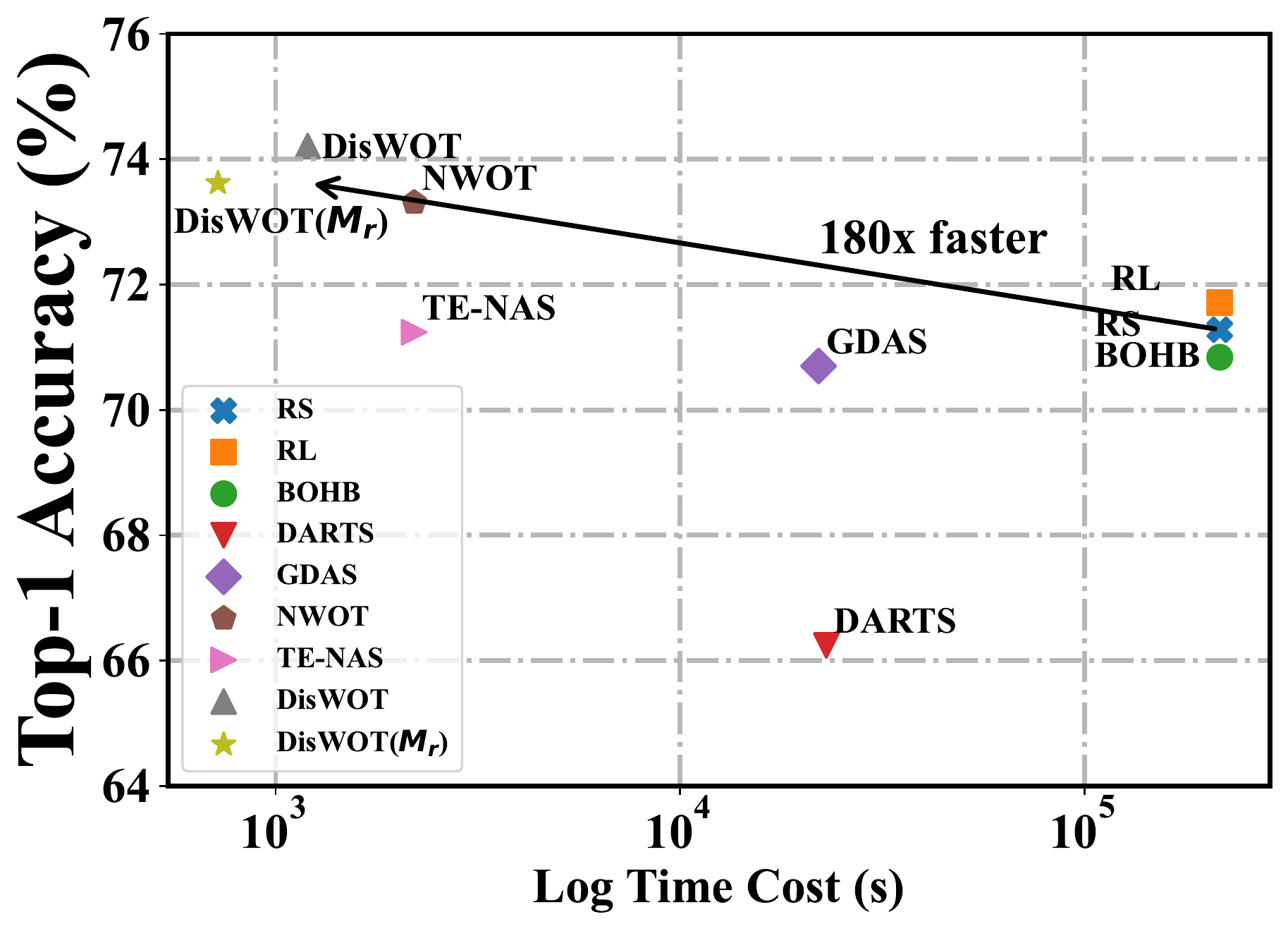}
  \end{minipage}
  \vspace{-0.3cm}
  \caption{Left: KD~\cite{ref10_kd}, DisWOT, DisWOT$\dag$ results for ResNet20 under different teachers. Right: Comparison of distill accuracy \& training efficiency with other NAS methods on NAS-Bench-201.}  
  
  \label{fig:2}
\end{figure}

Despite the remarkable achievements of Deep Neural Networks (DNNs) in numerous visual recognition tasks~\cite{li2021seg,wei2022convformer,wang2022mvsnet,wang2023flora,wangdionysus,wang2023ftso}, they usually lead to heavy costs of memory, computation, and power at model inference due to their large numbers of parameters. To address this issue, Knowledge Distillation (KD) has been proposed as a means of transferring knowledge from a high-capacity teacher model to a low-capacity target student model, providing a more optimal accuracy-efficiency trade-off during runtime~\cite{chen2022knowledge,zhao2022decoupled,beyer2022knowledge}. The original KD method~\cite{ref10_kd} utilizes the logit outputs of the teacher network as the source of knowledge. Subsequent studies~\cite{ref11_feature_kd, ref12_attention_kd, ref28_factor, ref43_vid, ref41_sp, AB, ref40_nst} have focused on extracting informative knowledge based on intermediate feature representations. However, as the gap in capacity between students and teachers increases, existing KD methods are unable to improve results, particularly in tasks that depend on large-scale visual models such as VIT and GPT-3~\cite{VIT,GPT-3}. For example, as shown in Figure~\ref{fig:2} (Left), the large teacher (e.g., ResNet110) lead to worse performance for the fixed student than the relatively smaller one (e.g., ResNet56).

\begin{figure}
\setlength{\abovecaptionskip}{0.cm}
\setlength{\belowcaptionskip}{-0.cm}
  \centering 
  \begin{minipage}[t]{0.49\linewidth}
    \centering
\includegraphics[width=1\linewidth]{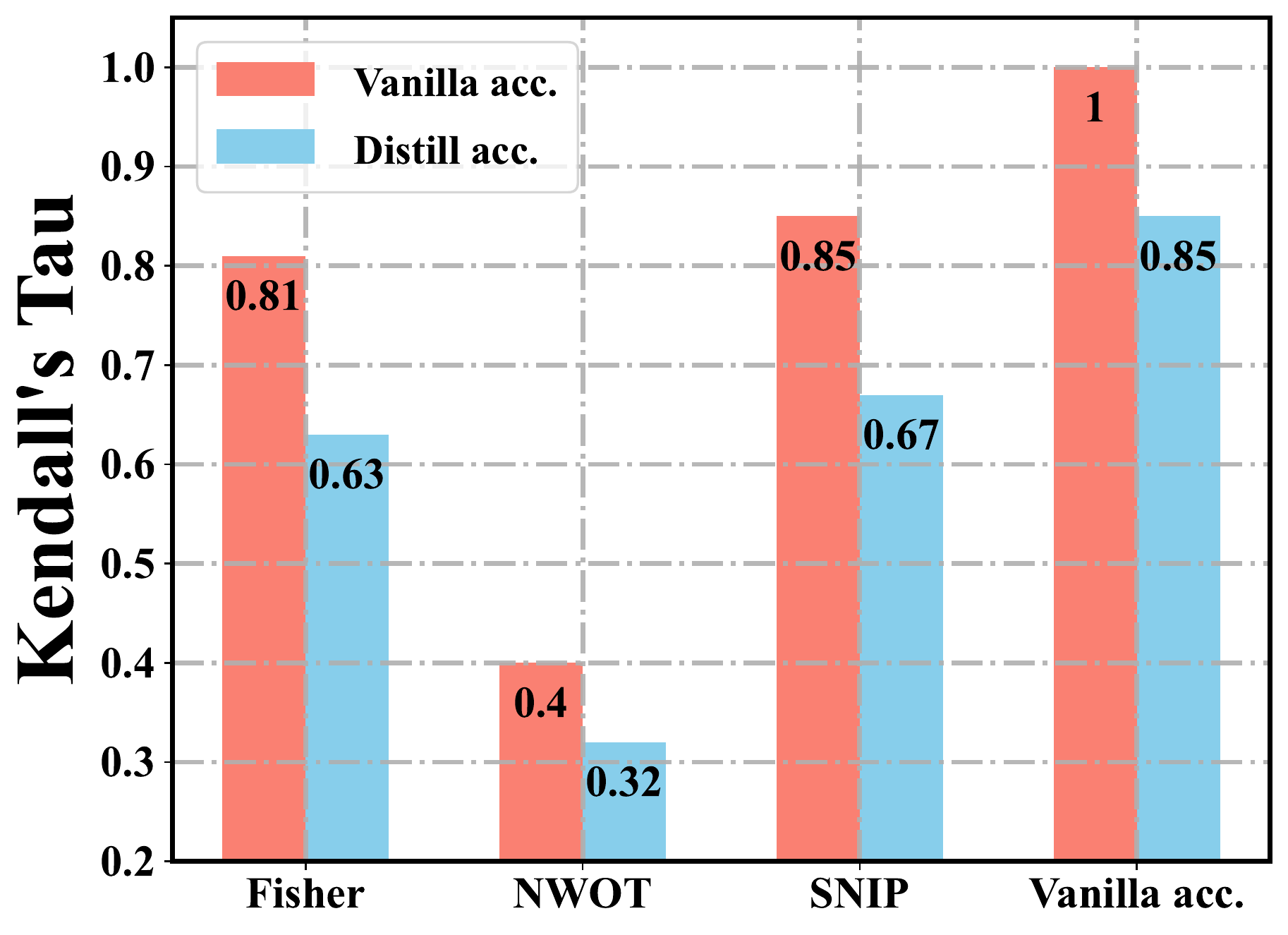}
  \end{minipage}%
  \begin{minipage}[t]{0.51\linewidth}
    \centering
\centering
\includegraphics[width=1\linewidth]{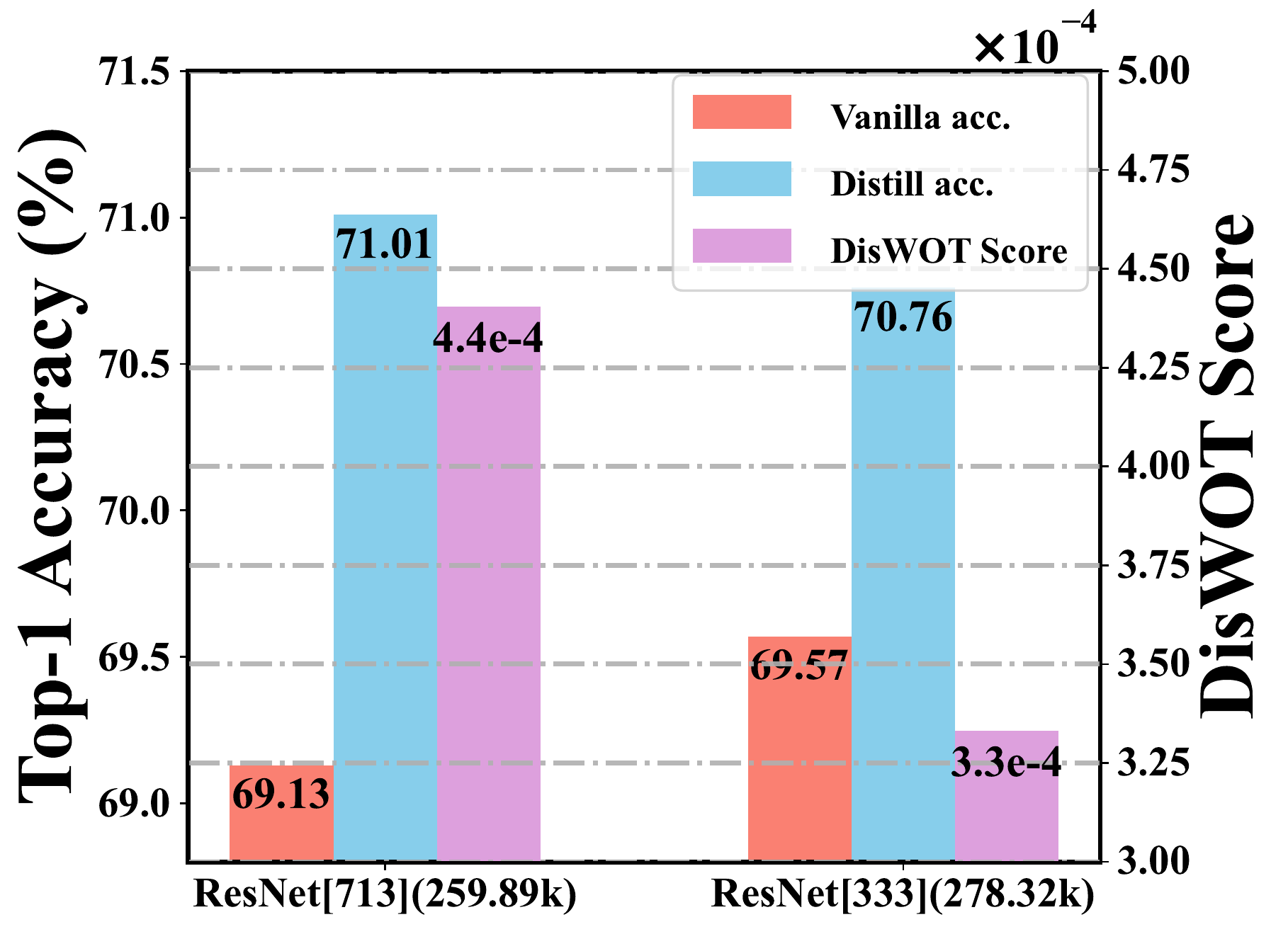}
  \end{minipage}
	\vspace{-0.3cm}
     \caption{Left:  Ranking correlation of proxies in zero-cost NAS with vanilla and distillation accuracy. Right: Vanilla accuracy, distillation accuracy, prediction scores of DisWOT for ResNet[7,1,3] and ResNet[3,3,3] on search space $S_0$.}  
  \label{fig:1}
\end{figure}

To solve this issue, adaptive KD methods have been proposed in terms of training paradigms~(e.g., early stop~\cite{cho2019efficacy}) and architectural adaptations~(e.g., assistant teacher~\cite{ATKD} and architecture search~\cite{ref22_search_student}), respectively. However, they are ineffective in improving distillation performance or involve enormous training costs in the additional model training and search process. In sharp contrast to these methods, we tackle this challenging problem from a new perspective regarding training-free architecture search.  To achieve this goal, we construct a search space  $S_0$ for ResNet-like models with different depth configurations and obtain vanilla and distill performance for each candidate in $S_0$ by individual training. Then, we evaluate the ranking correlation between predicted scores of training-free search methods and the actual performance of each student model. Surprisingly, as shown in Figure~\ref{fig:1} (Left), there are common ranking correlation loss ($10\%\downarrow\sim20\%\downarrow$) for these methods in predicting distillation accuracy than vanilla accuracy.  To clarify this,  we carefully analyze the disparities in vanilla and distillation performance for each model: (1) for overall search space,  vanilla accuracy only preserves  85\% correlations with actual distillation performance.  (2) for a particular instance, as shown in Figure~\ref{fig:1} (Right), ResNet20  with 3 res-blocks in each stage (i.e., ResNet[3,3,3]) has more parameters and better standalone performance but is weaker than ResNet[7,1,3] in the distillation process.  Considering that ResNet[7,1,3] has more layers than ResNet20, we seek to understand the above phenomenon regarding the vanilla-distillation accuracy gap from the perspective of semantic matching~\cite{lin2022knowledge}. ResNet[7,1,3] enjoys a larger effective receptive field and more excellent matched knowledge with teacher, resulting in significant distillation gains.  Encouraged by this understanding, we strive to design a new zero-proxy regarding the semantic matching of teacher-student.  As a result, we find that the similarity scores of feature semantics and sample relations can outperform conventional 
zero-cost NAS in predicting final distillation accuracy (see the comparison of ranking correlation on search space $S_0$ in Table~\ref{tab:zc_proxies}). As shown in Figure~\ref{fig:1}(Right), similarity scores are also consistent with distillation performance.

Drawing on the aforementioned observations, we introduce DisWOT, a simple yet effective training-free framework that finds the best student architectures for distilling the given teacher model. For better semantic matching in distillation, DisWOT leverages novel zero-cost metrics regarding the feature semantics and sample relations to select better student model. For the feature semantic similarity metric, we remark that randomly initialized models can localize objects well~\cite{random} and generate localization heatmaps via Grad-CAM~\cite{Grad} as reliable semantic information.  Then, we measure the channel-wise similarity matrix of localization heatmaps and take the $L_2$ distance of the similarity matrix for the teacher-student model as the metric.  For input samples, different models have diverse abilities to discriminate their relationships.  To improve relational knowledge matching ability, we use the $L_2$ distance of sample-relation correlation matrix as a relation similarity metric.  Finally, we search for student architectures using an evolutionary algorithm with semantic and relations similarity metrics.  Then, the distillation process is implemented between the searched student and the pre-defined teacher.  In addition, we leverage these metrics directly as new distillers to enhance the student, as the DisWOT$\dag$. Equipped with our train-free search and distillation design, our DisWOT and DisWOT$\dag$ framework significantly improve the model's accuracy-latency tradeoff in inference with at least 180$\times$ training acceleration.

In principle, our DisWOT use higher-order statistics of teacher-student models to optimize the student architecture to fit a given teacher model.  Its merits can be highlighted in three aspects: (1) In contrast to training-based student architecture search requires the individual or weight-sharing training, our DisWOT does not require the training of student models in the search phase.  In addition, DisWOT is efficient to compute and easy to implement as it uses only the mini-batch data at initialization.  (2) DisWOT is a teacher-aware search for distillation, which has better predictive distill accuracy than conventional NAS.  (3) DisWOT exploits the distance of higher-order knowledge between the neural networks, bridging knowledge distillation and zero-proxy NAS.  We further demonstrate the competitive ranking correlation of DisWOT among 10 knowledge distances in KD as zero-proxy for predicting vanilla accuracy in NAS-Bench-201.  We anticipate that our work on KD-based zero-proxy can offer some assistance in furthering research endeavors related to KD and NAS.

We conduct extensive experiments on CIFAR-100, ImageNet, and the NAS-Bench-201~\cite{dong2019search} dataset, demonstrating the superiority of our proposed approach.  In contrast to experiments in traditional architectural search, we focus on  final distillation accuracy instead of the vanilla accuracy for the student.  The results show that our DisWOT can achieve better accuracy than traditional Zero-shot NAS in the same search space.  Besides, by switching to a larger space, our DisWOT can obtain new state-of-the-art architectures.  For example, in the same ResNet-like search space, we significantly improved 1.62\% Top-1 accuracy over KD for ResNet50-ResNet18 pair under the same training settings.  We also conducted comprehensive ablation studies to investigate how our method can use the predictability of zero-cost metrics to boost the distillation performance. 

\noindent\textbf{Main Contributions:}
\begin{itemize}
\item By analyzing and exploring the discrepancy between teacher-student capability, we empirically show that their semantic similarities have a stronger correlation with the final distillation accuracy.  This motivates us to propose a new student architecture search for the Distillation without Training (DisWOT) framework to reduce the teacher-student capability gap, which, to the best of our knowledge, is not achieved in the area of knowledge distillation.

\item DisWOT proposes novel zero-cost metrics on similarity of feature semantics and sample relations and ensemble these metrics to select the optimal student via an evolutionary algorithm at the initial time. In the distillation stage,  DisWOT achieves state-of-the-art performances in multiple datasets and search spaces.

\item We further expand 10 kinds of knowledge distances including DisWOT as new universal KD-based zero proxies, which enjoy competitive predictive power with actual performance of models. We hope that our contributions in this endeavor may aid to some degree in advancing future research on KD and NAS. 

\end{itemize}

\section{Related Work and Background}
In this section, we summarize existing knowledge distillation and architecture search methods and clarify their differences to our method.

\subsection{General Formulation of Knowledge Distillation}
\label{sec: review}
The fundamental concept underlying Knowledge Distillation (KD) involves utilizing acquired knowledge (\eg, logits~\cite{lishadow}, feature values~\cite{li2020explicit,li2022self,li2022tf,li2022SFF,li2022norm}, and sample relations~\cite{ref13_rkd,ref14_relation_crd}) from a high-capacity teacher to guide the training of a student model. The training dataset $(X, Y)$ comprises training samples $X={x_i}_{i=1}^n$ and their corresponding labels $Y={y_i}_{i=1}^n$. Let $f_T$ be the output logits of the fixed teacher $T$ and let $f_S$ be the output of student $S$, respectively. In KD, the student network $f_S$ is trained by minimizing:
\begin{equation}
\label{eq:01}
\resizebox{0.90\hsize}{!}{
$\mathcal{L}_{S} =   \mathcal{L}_{CE}(f_S, Y)  + \mathcal{L}_{KL}\left(f_S, f_T\right)+  \mathcal{D}_{f}\big(\phi_S(x), \phi_T(x)\big)$,
}
\end{equation}
where  $\mathcal{L}_{CE}$ is the regular cross-entropy loss. $\mathcal{L}_{KL}$ represents Kullback-Leibler~(KL) divergence. $\mathcal{D}_{f}(\cdot,\cdot)$ is the distance function measuring the difference of intermediate feature representations (see Table~\ref{tab:FKD} for particular distillers).

\begin{table}[htbp]
  \centering
      \vspace{-0.1in} 
    \caption{Comparison of recent distillers.}
    \vspace{-0.1in} 
      \resizebox{0.95\linewidth}{!}{
    \begin{tabular}{llc}
    \toprule
    Method & Knowledge & Distance $\mathcal{D}_{f}(\cdot,\cdot)$ \\
    \midrule
    FitNets~\cite{ref11_feature_kd}  & Feature representation & $\mathcal{L}_{2}$ \\
    AT~\cite{ref12_attention_kd}    & Attention maps & $\mathcal{L}_{2}$ \\
    CC~\cite{ref40_cc}    & Instance relation &$\mathcal{L}_{2}$\\
    NST~\cite{ref40_nst}   & Neuron selectivity patterns & $\mathcal{L}_{MMD}$ \\
    PKT~\cite{ref42_pkt}   & Similarity probability distribution & $\mathcal{L}_{KL}$ \\
    \bottomrule
    \end{tabular}%
    }
     \label{tab:FKD}%
\end{table}%

\noindent\textbf{Comparison with Other Adaptive KDs for Distillation Gap.} DisWOT is the first train-free architecture search solution to reduce the teacher-student gap. Unlike training manners~\cite{cho2019efficacy} and KD-loss designs~, DisWOT utilizes the classic KD training configurations and distillers. In addition, DisWOT is free from the assistant teacher in ATKD~\cite{ATKD}, which involves a complex training routine and budget. AKD~\cite{ref22_search_student} searches student via reinforcement learning based on feedback from individual training of lots of models. As a completely alternative technical route to these training-based NAS~\cite{gu}, our training-free DisWOT builds on new zero-proxy and achieves 180$\times$ $\sim$ 1000$\times$ training acceleration, which greatly improves its easy-to-use and flexibility.

\begin{figure*}
\centering
\includegraphics[width=1\textwidth]{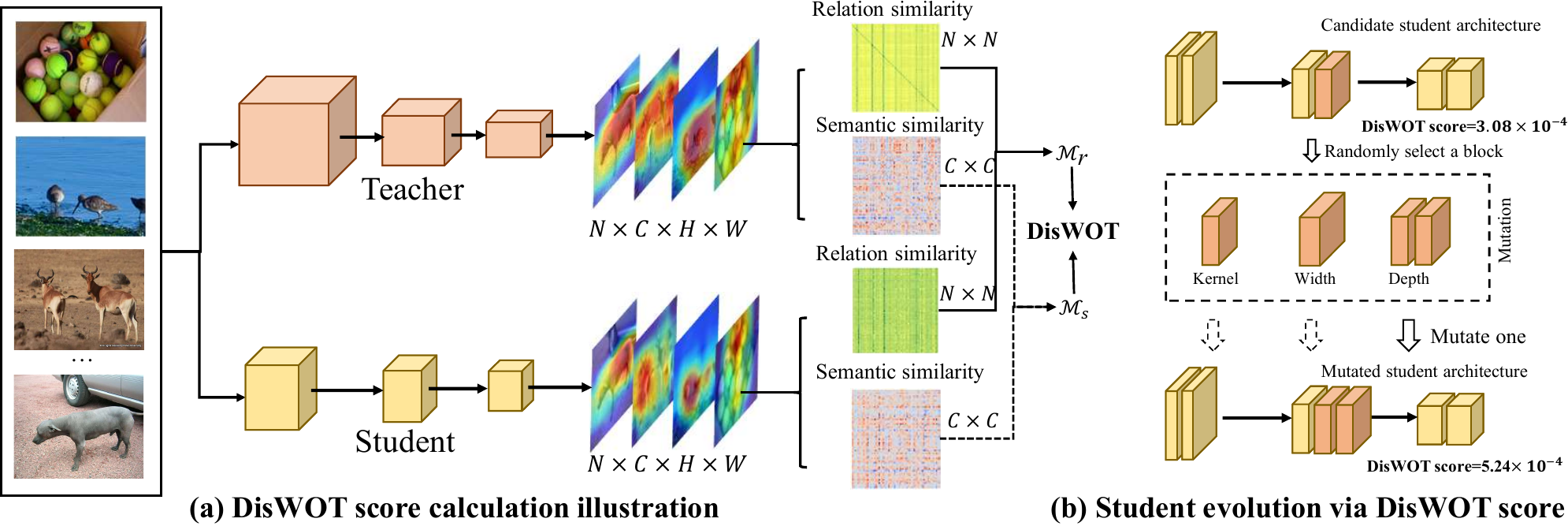}
  \caption{
A schematic overview of our DisWOT, including (a) detailed calculation of the DisWOT scores and (b) evolution of the student architecture via the DisWOT scores. In search phase, DisWOT use semantic similarity metrics and relations similarity metrics to select good student for a given teacher. The semantic similarity metric is measured by $l_2$ distance of the channel-wise correlation matrix for Grad-cam activation maps. Similarly, the relation similarity matrix statistics the sample-wise correlation matrix distance of the randomly initialized teacher-student pairs. With the feedback from these metrics, the evolutionary search in DisWOT automatically imitates good student from weak ones. In distillation phase, this searched student is distilled via  teacher model and achieves superior gains. }
\label{fig:overview}
    \vspace{-1em}
\end{figure*}

\begin{table}[]

    \centering
    \caption{Formulation of NAS methods. $\A$ is the search space. A candidate architecture in the search space is denoted as $\alpha \in \A$, which corresponds to a neural architecture $S(\alpha, w)$ with weight $w$. $\W$ is the weight of the supernet. $\Loss_{train}$ and $\Loss_{val}$ are the loss functions on the training and validation sets, respectively. }
    \resizebox{0.95\linewidth}{!}{
        \begin{tabular}{c|c|c}
            \hline
            Type                                       & Evaluation                & Formula                                   \\ \hline
            \multirow{8}{*}{\begin{tabular}[c]{@{}c@{}}Training- \\ based\end{tabular}} & \begin{tabular}[c]{@{}c@{}}Multi-trial\\ Training\end{tabular} & \parbox{4.5cm}{\begin{align}
                    \begin{split}
                        &\alpha^* = \arg\min_{\alpha \in \A} \Loss_{val}(S(\alpha, w_\alpha)), \label{equ:nas1} \\
                        &\text{s.t.}~ w_\alpha = \arg\min_{w} \Loss_{train}(S(\alpha, w))\nonumber
                    \end{split}
                \end{align}
            }                                                                                                                  \\ \cline{2-3}
                                                       & \begin{tabular}[c]{@{}c@{}}Weight\\ sharing\end{tabular} & \parbox{4.5cm}{\begin{align}
                    \begin{split}
                        &\alpha^* = \arg\min_{\alpha \in \A} \Loss_{val}(S(\alpha, \W_\A(\alpha))), \\
                        &\text{s.t.}~ \W_\A=\arg\min_{\W} \Loss_{train}(S(\A, \W))\nonumber 
                    \end{split}
                \end{align}
            }                                                                                                                  \\ \hline
            \begin{tabular}[c]{@{}c@{}}Training-\\ free \end{tabular}                  & \begin{tabular}[c]{@{}c@{}}Zero-cost\\ Proxy\end{tabular} & \parbox{5.0cm}{\begin{align}
                    \begin{split}
                        &\alpha^* =  \underset{\alpha \in \A} {ZeroProxy}(S(\alpha, w)) \nonumber
                    \end{split}
                \end{align}
            }                                                                                                                  \\ \hline
        \end{tabular}}
        \label{nas}
\end{table}

\begin{table}[htbp]
  \centering
  \caption{Comparison with different training-free NAS.}
  \resizebox{1\linewidth}{!}{
    \begin{tabular}{llcl}
    \toprule
    Type & Method & \multicolumn{1}{l}{Teacher-aware} & Objective \\
    \midrule
    Prune-based & SNIP~\cite{snip}, Fisher~\cite{zeronas}, Synflow~\cite{syflow} &   \xmark    & Vanilla acc. \\
    Activation-based & NWOT~\cite{NWOT}, Zen-NAS~\cite{ZenNAS} &   \xmark    & Vanilla acc. \\
    KD-based & DisWOT~(ours) &   \cmark    & Distill acc. \\
    \bottomrule
    \end{tabular}%
    }
  \label{tab:train-free-NAS}%
\end{table}%

\subsection{Revisiting Architecture Search Methods}
Neural Architecture Search (NAS) is emerged to reduce human efforts in architecture design and automate the discovery of high-performance networks. As formalized in Tab.~\ref{nas}, Multi-trial NAS methods~\cite{nasnet,ref22_search_student} train a large number of candidates individually, which leads to extensive resource consumption. To alleviate this,  many NAS~\cite{ENAS,li2021nas,lichengp,linas2} methods adopt a weight-sharing strategy within a single supernet to facilitate the simultaneous training of candidates. The supernet is trained for hundred of epochs by path sampling \cite{guo2019single,dong2023rd} or compound optimization with architecture representations \cite{liu2018darts,wu2019fbnet}. As an orthogonal direction, zero-cost NAS methods~\cite{NWOT,knas} focus on identifying well-performed architectures with training-free metrics. For example, NWOT~\cite{NWOT} calculates the architecture score based on the kernel matrix of binary activations between small batches of samples. 

\noindent\textbf{Comparison with NAS with Teacher.} Some training-based NAS~\cite{Zheng2022NeuralAS,peng2020cream,Li2020BlockWiselySN} employ a teacher model to supervise supernet training to improve predictive ability in the search stage. However, these methods aim to improve vanilla accuracy, not for distillation, and they do not use the teacher model in the full training stage. In addition, without any training costs, our training-free DisWOT enjoys obvious differences than these methods and advantages in efficiency.

\noindent\textbf{Compared to Other Training-free NAS.} Table~\ref{tab:train-free-NAS} clearly summarizes the differences between DisWOT and other zero-cost methods\cite{zeronas,ZenNAS,NWOT,snip,syflow}. Moreover, DisWOT outperforms these methods on distillation performance prediction and boosting in our sufficient experiments dealing with diverse datasets and search spaces.

\section{Methodology}
Figure \ref{fig:overview} provides an overview of the \ourmethod{} framework, which is comprised of two main stages: optimal student network search and distillation with high-order knowledge. In the search stage,  we employ the neural architecture search technique to obtain an optimal student network for a pre-defined teacher network. Notably, we propose a training-free proxy called DisWOT to accurately rank enormous student networks and prevent expensive evaluation processes with high efficiency.
In the distillation stage, the searched student network is retrained with distillation to imitate high-order knowledge in the teacher network. We give the details of these two designs in the following sections.

\subsection{Search for Optimal Student Network}
We first present the training-free metrics we designed to score a student architecture, which indicates its final accuracy when distilled with a pre-defined teacher network. Then we depict the details of the evolutionary process to obtain an optimal student candidate.

\noindent\textbf{Semantic Similarity Metric.} 
The semantic information is meaningful for neural networks to perceive as humans. In distillation, the teacher network always has more convolutional operations than the student, resulting in a teacher feature map with a larger receptive field and greater richness of semantic information. In contrast to distiller designs to alleviate semantic gaps, we aim for train-free student architecture to better match the teacher model with computational constraints. We notice that the network with random initial weights also has some semantic localization capability. Thus, we start to analyze the localization performance of the randomly initialized teacher-student model. Specifically, we utilize  Grad-CAM maps~\cite{zhou2016learning} to localize semantic object regions, which explains the model decisions using gradient information.  Given a mini-batch of input images, we define the high-level feature map before the Global Average Pooling (GAP) layer of the teacher network $T$ as $A_T\in \mathbf{R}^{B\times C_T\times H_T\times W_T}$, where $B$ represents the batch size, $C_T$ denotes the number of output channels, and $H_T$ and $W_T$ are the spatial dimensions. Additionally, we introduce $A_T^c \in \mathbf{R}^{N\times H_T\times W_T}$ as the c-th spatial map along the channel dimension. For the student network $S_i$, we have feature map $A_{S_i}^c \in \mathbf{R}^{B\times C_S\times H_S\times W_S}$ and spatial map $A_{S_i}^c \in \mathbf{R}^{B\times H_S\times W_S}$, respectively. To compute the Grad-CAM maps of the $n$-th class for both the teacher and student networks, we can use the following formulations:

\begin{small}
\begin{equation}
	G_T=\sum_{c=1}^{C_T} w_{n,c}^T{A_T^c}, \quad
		G_{S_i}=\sum_{c=1}^{C_S} w_{n,c}^S{A_{S_i}^c},
\end{equation}
\end{small}where  $w^T \in \mathbf{R}^{N\times C_T}$ and $w^S \in \mathbf{R}^{N\times C_S}$ are weights of the last fully-connected layer in the teacher and student network.
$N$ represents the number of classes.
$w_{n,c}^T$ and $w_{n,c}^S$ refer to the element located in the $n$-th row and $c$-th column of weight matrices $w^T$ and $w^S$, respectively. To quantify the intersection of class-discriminative localization maps, we formulate 
semantic similarity metric $\mathcal{M}_{s}$ as the inter-correlation on the accumulated Grad-CAM maps for both teacher and student networks as follows:
\begin{align}
    \mathcal{G}^{T}=\frac{(G_T)\cdot(G_T)^\top}{\left\| (G_T)\cdot(G_T)^\top\right\| _2},   \mathcal{G}^{S}=\frac{(G_S)\cdot(G_S)^\top}{\left\| (G_S)\cdot(G_S)^\top\right\| _2},
    \label{eq:eq3}
\end{align}

\begin{equation}
\label{sample_metric}
    \mathcal{M}_{s}=\left\|\mathcal{G}^{T}- \mathcal{G}^{S_i}\right\| _2.
\end{equation}

\noindent\textbf{Relation Similarity Metric.} The relationships between input samples are non-trivial for knowledge transfer. To reduce the teacher-student gap and improve the relation-distillation performance, we use the correlation matrix as the sample-wise metric to search for an optimal student network. For the random teacher network $T$ and student network $S_i$ with activation maps $A_T\in \mathbf{R}^{N\times C_T\times H_T\times W_T}$ and $A_{S}^{i}\in \mathbf{R}^{N\times C_i\times H_i\times W_i}$, the correlation matrix of the mini-batch samples in the teacher network is formulated as follows:
\begin{align}
 \mathcal{A}^{T}=\frac{(\tilde{A}_T)\cdot (\tilde{A}_{T})^{\,\,\top}}{\left\| (\tilde{A}_T)\cdot (\tilde{A}_{T}^{\,\,\top})\right\| _2} , \mathcal{A}^{S_i}=\frac{(\tilde{A}_S)\cdot (\tilde{A}_{S})^{\,\,\top}}{\left\| (\tilde{A}_S)\cdot (\tilde{A}_{S}^{\,\,\top})\right\| _2} ,
\end{align}
\noindent where $\tilde{A}_T \in \mathbf{R}^{N \times CHW}$ is a reshaping of $A_T$, and $M_T$ is a $N \times N$ matrix. Thus, the $(i,j)$ entry in matrix $C_T$ represents the similarity between the $i$-th and $j$-th images within the mini-batch. Based on this, the sample similarity metric $\mathcal{M}_{r}$ for a potential student model $S_i$ is defined as follows:
\begin{equation}
    \mathcal{M}_{r}=\left\|\mathcal{A}^{T}- \mathcal{A}^{S_i}\right\| _2.
\end{equation}\label{similarity_metric}

\noindent\textbf{Training-Free Evolutionary Search.} 
Based on the above metric, we conduct a training-free evolutionary search algorithm to efficiently discover the optimal student $\alpha^* $ from  search space $\A$, as: 
\begin{equation}
\alpha^* =  \arg\min_{\alpha \in \A}(\mathcal{M}_{s}+\mathcal{M}_{r}).
\end{equation} 
\noindent \textbf{Theoretical Understanding.} According to the VC theory ~\cite{vapnik1998statistical}, the classification error of the vanilla teacher-student network can be decomposed as follows:
\begin{equation}
\resizebox{0.9\hsize}{!}{
    $R(f_s) - R(f_r) \le O \left(
    \frac{|\mathcal{F}_s|_C}{n^{\alpha_{sr}}} \right) + \epsilon_{sr}; R(f_t) - R(f_r) \le O \left(
    \frac{|\mathcal{F}_t|_C}{n^{\alpha_{tr}}} \right) + \epsilon_{tr},$}
\end{equation}\label{eq:teacher-from-real}
where $f_s \in \mathcal{F}_s$ is the student function, $f_t \in \mathcal{F}_t$ is the teacher function, and $f_r \in \mathcal{F}_r$ is the target function.
$R$ is the error.
$O(\cdot)$ and  $\epsilon_{sr}$ terms are the estimation and approximation error, respectively. 
$O(\cdot)$ is related to the statistical procedure when given the number of data points. In contrast, $\epsilon_{sr}$ is the approximation error of the student function class $\mathcal{F}_s$ for $f_r \in \mathcal{F}_r$. $|\cdot|_C$ is an function class capacity measure, and $n$ is the number of data point. During distillation, the student network is supervised purely with the teacher network as follows: 
\begin{equation}
    R(f_s) - R(f_t) \le O \left(
    \frac{|\mathcal{F}_s|_C}{n^{\alpha_{st}}} \right) + \epsilon_{st},
    \label{eq:student-from-teacher}
\end{equation}
where $\alpha_{st}$ and $ \epsilon_{st}$ are associated to student learning from teacher.
By combining Equations \ref{eq:teacher-from-real} and \ref{eq:student-from-teacher}, we obtain:
\begin{equation}
 \resizebox{0.88\hsize}{!}{
    $R(f_s) - R(f_r) \le O \left(
    \frac{|\mathcal{F}_t|_C}{n^{\alpha_{tr}}} \right) + \epsilon_{tr} + O \left(
    \frac{|\mathcal{F}_s|_C}{n^{\alpha_{st}}} \right) + \epsilon_{st}.$
}    
\end{equation}\label{eq:inequal} When student obtains gains in KDs, its upper bound of error in distillation is smaller than vanilla training, which satisfies the following inequality:
\begin{equation}
    \resizebox{0.85\hsize}{!}{$O \left(
    \frac{|\mathcal{F}_t|_C}{n^{\alpha_{tr}}}
    + \frac{|\mathcal{F}_s|_C}{n^{\alpha_{st}}}
    \right) + \epsilon_{tr} + \epsilon_{st}  \le O \left(
    \frac{|\mathcal{F}_s|_C}{n^{\alpha_{sr}}} \right) + \epsilon_{sr}.$}
\end{equation}\label{eq:blkd}
Based on the assumption in \cite{ref10_kd} that $\epsilon_{tr} + \epsilon_{st} \le \epsilon_{sr}$ holds consistently, we focus on minimizing $ O \left(
    \frac{|\mathcal{F}_t|_C}{n^{\alpha_{tr}}}
    + \frac{|\mathcal{F}_s|_C}{n^{\alpha_{st}}}
    \right)$ to improve the distillation performance. As noted in Lopez-Paz et al~\cite{lopez2015unifying}, a better representation allows for a faster learning rate with a fixed amount of data. Hence, when there is a larger gap between the capacities of the student and teacher networks, the value of $\alpha_{st}$ tends to be lower.
Thus we aim to search for an optimal student network that meets the requirement of $\alpha_{s_{i}t} \le \alpha_{s_{o}t}$, where $s_i$ is all candidate student networks, and $s_o$ is our searched student network. In this case, the inequality becomes more effective, and we improve the knowledge distillation by injecting a larger $\alpha_{s_{o}t}$. 
Specifically, we present the overall procedure for discovering optimal student in algorithm \ref{alg:evolution}.

\begin{figure}
\setlength{\abovecaptionskip}{0.cm}
\setlength{\belowcaptionskip}{-0.cm}
  \begin{minipage}[t]{0.51\linewidth}
    \centering
\includegraphics[width=1\linewidth]{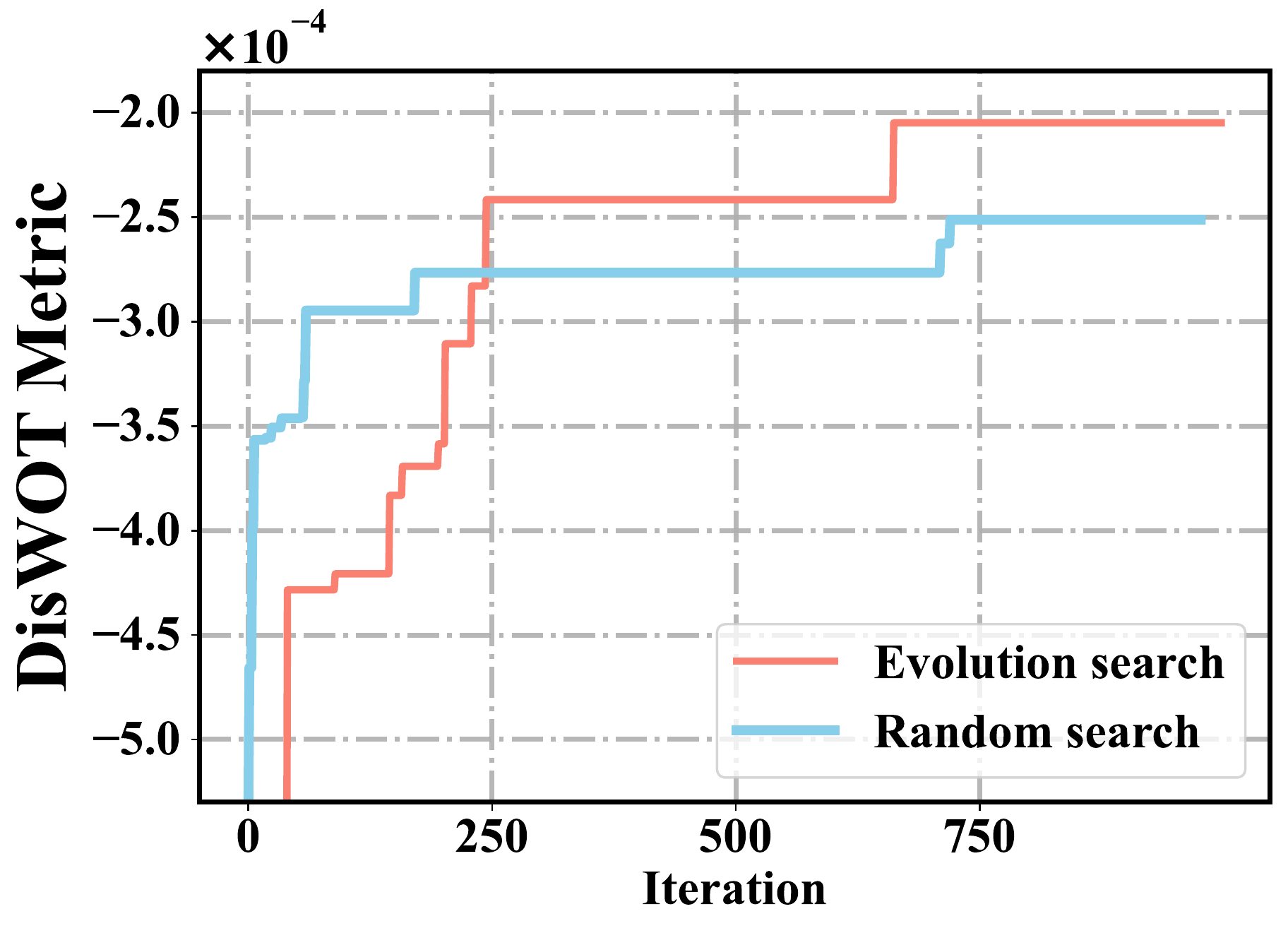}
  \end{minipage}%
  \begin{minipage}[t]{0.49\linewidth}
    \centering
\centering
\includegraphics[width=0.99\linewidth]{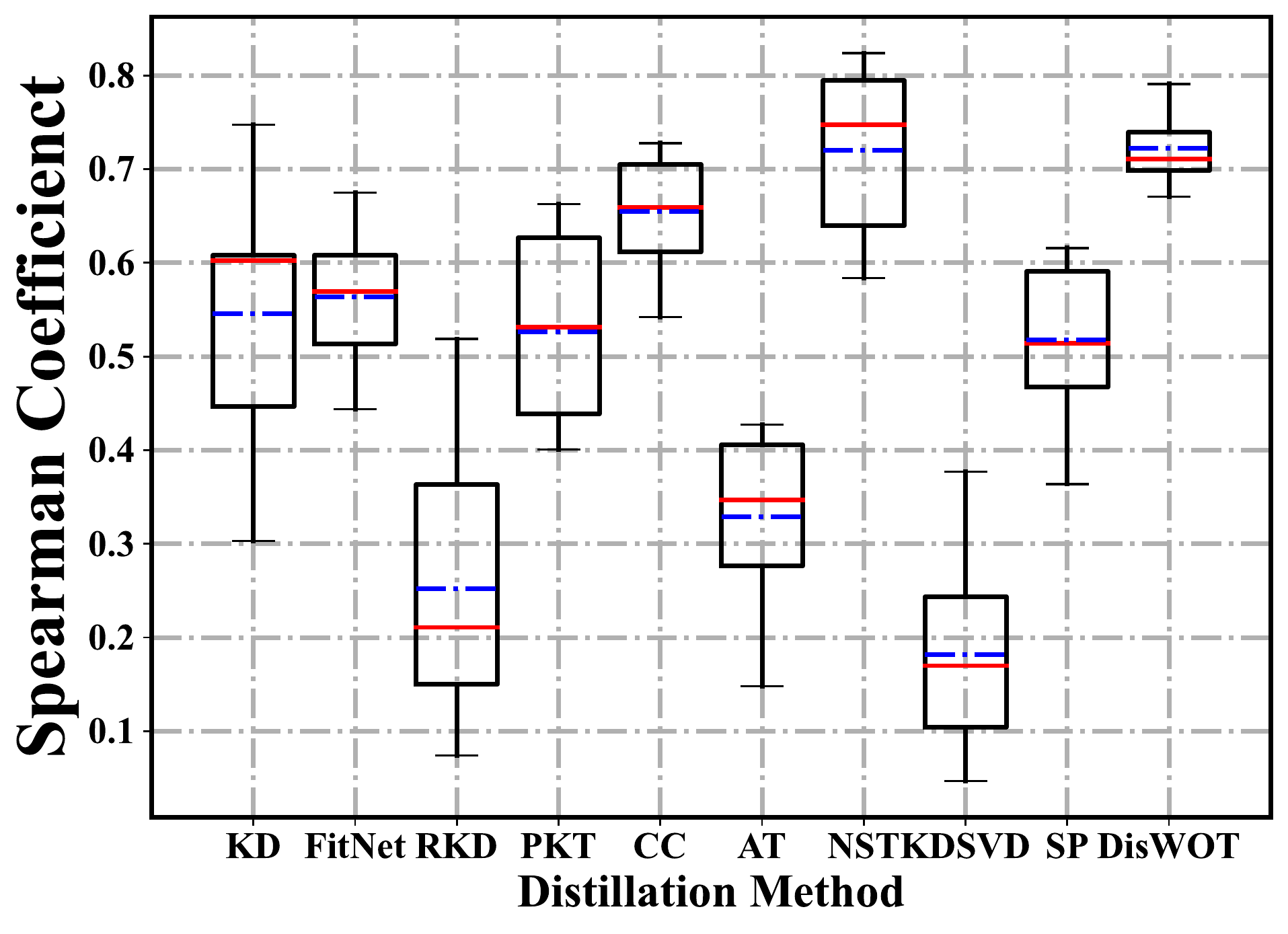}
  \end{minipage}
 \caption{Left: comparison of random search and evolution search. Right: ranking correlation of different distillation methods on NAS-Bench-201.}
  \label{fig:es_rs}
\end{figure}

\noindent\textbf{Effects of Search Strategies.} We compare the evolution search algorithm and the random search algorithm in search space $S_2$ with the same number of iterations, as shown in Figure~\ref{fig:es_rs}. We find that the evolution search algorithm can consistently find architectures with lower DisWOT, especially when the search space is relatively large, and the evolutionary search can explore better architectures.

\begin{algorithm}
\small
\caption{Evolution Search for DisWOT}
\label{alg:evolution}
\textbf{Input}: Search space $\mathcal{S}$, population $\mathcal{P}$, architecture constraints $\mathcal{C}$, max iteration $\mathcal{N}$, sample ratio $r$, sampled pool $\mathcal{Q}$, topk $k$, teacher network $\mathcal{T}$.

\textbf{Output}: \leftline{Highest DisWOT score architecture.}
\begin{algorithmic}[1]
\STATE $\mathcal{P}_0$ := Initialize population$(P_i, \mathcal{C}$); 
\STATE sample pool $\mathcal{Q}$ := $\emptyset$;
\FOR{$i = 1:\mathcal{N}$}
    \STATE Clear sample pool $\mathcal{Q}$ := $\emptyset$;
    \STATE Randomly select $r \times \mathcal{P}$ subnets $\hat{P_i} \in \mathcal{P}$ to get $\mathcal{Q}$;
    \STATE Candidates $\{A_i\}_{k}$ := GetTopk($\mathcal{Q}$, $k$);
    \STATE Parent $A_i$ := RandomSelect$(\{A_i\}_{k})$;
    \STATE Mutate $\hat{P_i}$ := MUTATE($A_i$);
    \IF{$\hat{P_i}$ do not meet the constraints $\mathcal{C}$} 
        \STATE Do nothing;
    \ELSE
        \STATE Get DisWOT-Score $z$ := DisWOT($\hat{P_i}$, $\mathcal{T}$);
        \STATE Append $\hat{P_i}$ to $\mathcal{P}$;
    \ENDIF 
    \STATE Remove network of smallest DisWOT-score;
\ENDFOR
\end{algorithmic}
\end{algorithm}

\subsection{Distillation with High-order Knowledge}
In the distillation stage, teacher model $T$ is employed to distill the optimal student network $f_S$. To verify the superiority of our search architecture, we adopt the existing distillers (e.g., KD) as the default distillation setting. In addition, we observe that the metrics we searched for actually serve as minimization optimization goals in the distillation process to transfer the teacher's privileged semantic and sample relational knowledge as the semantic distillation and sample distillation:
\begin{small}
\begin{equation}
      \mathcal{L}_{\mathcal{M}_{s}} = \frac{1}{c^2} \left\|\mathcal{G}^{T}- \mathcal{G}^{S}\right\| _2,    \mathcal{L}_{\mathcal{M}_{r}} = \frac{1}{b^2} \left\|\mathcal{A}^{T}- \mathcal{A}^{S}\right\| _2,
\end{equation}
\end{small}

Finally, we involve these advanced distillers in our framework, called DisWOT$\dag$. The total loss for DisWOT and DisWOT$\dag$ as:
\begin{equation}
\label{eq:diswot}
\begin{split}
\mathcal{L}_{\text{DisWOT}}= \mathcal{L}_{CE}(f_S, Y)  + \mathcal{L}_{KL}\left(f_S, f_T\right), \\ \mathcal{L}_{\text{DisWOT$\dag$}}=\mathcal{L}_{\text{DisWOT}}+\mathcal{L}_{\mathcal{M}_{s}}+\mathcal{L}_{\mathcal{M}_{r}}.
\end{split}
\end{equation}

\begin{table}[]
\centering
\small
\caption{Spearman correlation $\rho$ (\%) on NAS-Bench-201.}
  \resizebox{1\linewidth}{!}{
\begin{tabular}{clclc}

\hline
 Type         &      Method     & $\rho$ &     Method    & $\rho$ \\ \hline
\multirow{4}{*}{\begin{tabular}[c]{@{}c@{}}Zero-cost \\ Proxies\end{tabular}}    & Grad\_Norm~\cite{zeronas} & 58.70$\pm$0.11     & Synflow~\cite{syflow} & \textbf{74.61$\pm$0.08}     \\
                                      & SNIP~\cite{snip}      & 58.17$\pm$0.15     & Jacob~\cite{jacob}   & 73.42$\pm$0.03     \\
                                      & Fisher~\cite{zeronas}    & 35.91$\pm$0.09     & Zen-NAS~\cite{ZenNAS} & 41.36$\pm$0.06     \\
                                      & NWOT~\cite{NWOT}      & 64.41$\pm$0.08     & FLOPs~\cite{zeronas}   & 63.38$\pm$0.06     \\ \hline
\multirow{4}{*}{\begin{tabular}[c]{@{}c@{}}KD-based \\ Proxies\end{tabular}} & KD~\cite{ref10_kd}        & 54.43$\pm$0.09     & PKT~\cite{pkt}     & 52.65$\pm$0.09     \\
                                      & FitNets~\cite{fitnets}    & 56.18$\pm$0.09     & CC~\cite{cc}      & 65.90$\pm$0.08     \\
                                      & SP~\cite{sp}        & 51.24$\pm$0.08     & NST~\cite{nst}     & 72.35$\pm$0.09     \\
                                      & RKD~\cite{rkd}       & 25.71$\pm$0.17     & DisWOT  & \textbf{72.36$\pm$0.02}     \\ \hline
\end{tabular}\label{tab:zero-proxie}
}
\end{table}

\subsection{Bridging Distiller and Zero-proxy}
In our DisWOT framework, we use the semantic and relational similarity metrics as a distillation performance predictor and distiller. In addition, DisWOT also enjoys good performance for vanilla performance predictions. Encouraged by this intriguing observation, we employ the knowledge function in of different KDs as zero-proxies and evaluate their ranking consistency with vanilla accuracy. As shown in Table~\ref{tab:zero-proxie}, these 
KD-based zero-proxies enjoy competitive rankings with other NAS methods. Detailed results in Figure~\ref{fig:es_rs} illustrate that our DisWOT and NST~\cite{ref40_nst} are the winners in the family of KD-based proxies. These attempts reveal the close connections between KD and NAS, and augment 10+ new universal proxies from the teacher-student learning perspective for training-free NAS research.

\section{Experimental results}

In this section, we present the experimental results of our DisWOT on different datasets. First, we describe the four datasets used in our experiments and three search spaces $S_0,S_1,S_2$ in Sec. \ref{exp:setup}. Then, we conduct a comprehensive set of experiments to evaluate the effectiveness of \ourmethod{}.

\subsection{Experimental Setup}\label{exp:setup}
We perform experiments on four datasets, namely CIFAR-10, CIFAR-100, ImageNet-16-120, and ImageNet-1k. In the search process, we only use one batch of training data to get the statistic at nearly no cost. Following previous works, our  experiments are conducted on the following three search spaces:

\noindent \textbf{Search Space $S_0$:} Following cifar-ResNet\cite{he2016deep}, the search space consists of three residual blocks and is based on CIFAR-100 datasets. The depth of each residual block is searched in set \{1,3,5,7\}. 

\noindent \textbf{Search Space $S_1$:} Following NAS-Bench-201~\cite{dong2019search}, this Darts-like search space is a cell-based search space consisting of stacked directed acyclic graphs. $S_1$ is conducted on CIFAR-10, CIFAR-100, and ImageNet-16-120 datasets.

\noindent \textbf{Search Space $S_2$:} Following NDS\cite{Radosavovic2020DesigningND}, this search space consists of residual and bottleneck blocks defined in ResNet. $S_2$ is based on CIFAR-100 and ImageNet-1k dataset.

\subsection{Experiments on CIFAR-100}\label{exp:s0}

\noindent \textbf{Implementation Details.} We compare distillation gains with other zero-nas on search space $S_1$. In the search phase, we configure 48k evolution iters with 512 population sizes. In distillation, All searched student networks are trained via CRD's settings~\cite{ref14_relation_crd} with ResNet56 as the teacher model.

\begin{table}[htbp]
  \centering
 \caption{Distillation results (\%) of different zero-cost proxies with knowledge distillation methods under 1M parameters. }
   \resizebox{1\linewidth}{!}{
    \begin{tabular}{l|lllll}
    \toprule
    Method & Random & FLOPs & Synflow & NWOT  & DisWOT \\
    \midrule
    Baseline & 69.52 & 71.37 & 72.88 & 71.80 & 73.12 \\
    KD    & 70.45 & 72.13 & 73.72 & 72.57 & 74.73 \\
    FitNets & 70.12 & 72.40 & 73.55 & 72.72 & 74.85 \\
    AT    & 70.16 & 72.97 & 73.52 & 72.08 & 74.50 \\
    SP    & 70.46 & 72.14 & 73.50 & 72.16 & 74.95 \\
    RKD   & 71.19 & 72.22 & 73.69 & 72.63 & 74.62 \\
    CRD   & 71.59 & 72.78 & 73.99 & 73.12 & 75.25 \\
    \bottomrule
    \end{tabular}%
    }
  \label{tab:under_1M}%
\end{table}%

\noindent \textbf{Distillation Results of Zero-cost Proxies.} We conduct detailed experiments on other zero-cost proxies with different knowledge distillation methods. Note that we search the student network under constraints of 1M parameters. The results in Table~\ref{tab:under_1M} demonstrated that our proposed DisWOT achieved superior results compared with other zero-cost proxies with different knowledge distillation methods. The DisWOT outperforms its counterparts vanilla networks by around 2\%, while achieving consistent improvements among different distillation methods, such as KD~\cite{ref10_kd}, FitNets~\cite{fitnets}, AT~\cite{at}, SP~\cite{sp}, RKD~\cite{rkd}, and CRD\cite{crd}.

\begin{table}[htbp]
  \centering
  \caption{Distillation results(\%) of zero-cost proxies under \{0.5,1,2\}M parameters. }
    \resizebox{0.99\linewidth}{!}{
    \begin{tabular}{l|llll}
    \toprule
    Param.   & FLOPs & NWOT  & DisWOT & DisWOT$\dag$ \\
    \midrule
    0.5M   & 69.88 & 70.38 & 72.89 & \textbf{73.75} \\
    1M     & 72.13 & 72.57 & 74.23 & \textbf{75.25} \\
    2M     & 73.27 & 73.86 & 75.95 & \textbf{76.67} \\
    \bottomrule
    \end{tabular}%
    }
  \label{tab:diff_param_exp1}%
\end{table}%

\begin{table}[htbp]
    \centering
    \small
  \caption{Distillation results(\%) of zero-cost proxies under \{50,100\}M FLOPs on space $S_1$. }
    \resizebox{1\linewidth}{!}{
    \begin{tabular}{l|lll|l|lll}
    \toprule
    FLOPs  & NWOT  & Synflow & DisWOT & FLOPs  & NWOT  & Synflow & DisWOT \\
    \midrule
    50M   & 63.19 & 64.28 & 65.98 & 100M  & 70.38 & 72.12 & 72.89 \\
    \bottomrule
    \end{tabular}
    }
    \label{tab:diff_param_exp2}
\end{table}

\begin{table}[htbp]
  \centering
  \small
    \caption{Ranking correlation~(\%) of zero-cost proxies on $S_0$ space on CIFAR-100. }
  \resizebox{1\linewidth}{!}{
    \begin{tabular}{lccc}
    \toprule
    Method & Kendall's Tau    & Spearman     & Pearson\\
    \midrule
    FLOPs~\cite{zeronas} & 51.61 & 72.92 & 76.40 \\
    Fisher~\cite{zeronas} & 62.86 & 81.37 & 20.90 \\
    Grad\_Norm~\cite{zeronas} & 63.75 & 82.35 & 39.35 \\
    SNIP~\cite{snip}  & 67.22 & 85.07 & 51.09 \\
    NWOT~\cite{NWOT}  & 31.87 & 45.66 & 48.99 \\
    DisWOT (ours) & \textbf{73.98} & \textbf{91.38} & \textbf{84.83} \\
    \bottomrule
    \end{tabular}%
    }
  \label{tab:zc_proxies}%
\end{table}%

\noindent \textbf{Analysis on Varying Parameter Constraints.} We analyze the performance of student models under different parameter constraints obtained by DisWOT on CIFAR-100. 
As shown in the Table~\ref{tab:diff_param_exp1}, we compared our method with two zero proxies, a.k.a. FLOPs~\cite{zeronas} and NWOT~\cite{NWOT}, under the parameter constraints of 0.5, 1, and 2M, respectively, and the results demonstrate that our method still achieves excellent results.  As shown in Tab.~\ref{tab:diff_param_exp2}, DisWOT also outperforms previous SOTA methods with 0.8\%$\sim$1.7\%$\uparrow$ gains under same FLOPs constraints,

\begin{table*}[t]
	\caption{Distillation results on CIFAR-10, CIFAR-100, and ImageNet-16 in NAS-Bench-201~\cite{nas201}. Dis. Acc.~(\%)  represents the accuracy of the searched architecture after distillation training.  Time~(s) denotes the time cost (GPU-seconds) during the search phase.  The results of NWOT and TE-NAS come from their original papers. Our DisWOT achieves competitive results with the lowest costs.
	}
	\label{tab:nas}
	\centering
	\footnotesize
	 \resizebox{1\linewidth}{!}{
	\begin{tabular}{c|c|crr|crr|crr}
		\toprule
		\multirow{2}{*}{Type}           & \multirow{2}{*}{Model} & \multicolumn{3}{c|}{CIFAR-10} & \multicolumn{3}{c|}{CIFAR-100} & \multicolumn{3}{c}{ImageNet-16-120}                                                                \\
		\cline{3-11}
		                                &                        & Dis. Acc(\%)                     & Time~(s)                       & Speed-up                        & Dis.Acc(\%) & Time (s) & Speed-up & Dis. Acc(\%) & Time~(s) & Speed-up \\

		\midrule
		\multirow{5}{*}{Multi-trial}        & RS                     & 93.63                        & 216K                          & 1.0$\times$                            & 71.28   & 460K    & 1.0$\times$     & 44.88    & 1M      & 1.0$\times$     \\
		                                & RL~\cite{baker2016designing}                     & 92.83                        & 216K                          & 1.0$\times$                            & 71.71   & 460K    & 1.0$\times$     & 44.35    & 1M      & 1.0$\times$     \\
		                                & BOHB~\cite{Falkner2018BOHBRA}                   & 93.49                        & 216K                          & 1.0$\times$                            & 70.84   & 460K    & 1.0$\times$     & 44.33    & 1M      & 1.0$\times$     \\
		                                & RSPS~\cite{Li2019RandomSA}                   & 91.67                        & 10K                           & 21.6$\times$                           & 57.99   & 46K     & 21.6$\times$    & 36.87    & 104K    & 9.6$\times$     \\
		\midrule
		\multirow{2}{*} {Weight-sharing}      & GDAS~\cite{GDAS}                   & 93.39                        & 22K                           & 12.0$\times$                           & 70.70   & 39K     & 11.7$\times$    & 42.35    & 130K    & 7.7$\times$     \\
		                                & DARTS~\cite{darts}                  & 89.22                        & 23K                           & 9.4$\times$                            & 66.24   & 80K     & 5.8$\times$     & 43.18    & 110K    & 9.1$\times$     \\
		\midrule
		\multirow{2}{*} {Training-free} & NWOT~\cite{NWOT}                 & 93.73                        & 2.2K                          & 100$\times$                            & 73.31   & 4.6K    & 100$\times$     & 45.43    & 10K     & 100$\times$     \\
		                                & TE-NAS~\cite{TENAS}                 & 93.92                        & 2.2K                          & 100$\times$                            & 71.24   & 4.6K    & 100$\times$     & 44.38    & 10K     & 100$\times$     \\
		\midrule

		\multirow{2}{*} {DisWOT}        & \textbf{$\mathcal{M}_{s}$ \& $\mathcal{M}_{r}$}            & 93.55                        & 1.2K                          & 180$\times$                            & 74.21   & 9.2K    & 180$\times$     & 47.30    & 20K     & 180$\times$     \\
		                                & \textbf{$\mathcal{M}_{r}$}               & 93.49                        & 0.72K                         & \textbf{300$\times$}                            & 73.62   & 18.4K   & \textbf{300$\times$}     & 45.63    & 40K     & \textbf{300$\times$}     \\
		\bottomrule
	\end{tabular}
	}
	\label{tab:nb201_sota}
\end{table*}

\noindent\textbf{Ranking Correlation with Distill Accuracy.} Based on search space $S_0$, we perform vanilla training and distillation for each candidate with CRD's settings~\cite{ref14_relation_crd}. Then, we collect these  vanilla results as GT and analyze the different zero-proxy's correlation with them. As shown in Table~\ref{tab:zc_proxies}, the results illustrate that our DisWOT achieves higher than Fisher, GradNorm, SNIP,FLOPs, and NWOT by a large margin, and achieve results that are on par with the best zero-cost proxy, a.k.a. Zen-NAS and Synflow, on Kendall's Tau, Pearson, and Spearman coefficient.

\begin{table*}[htbp]
  \centering
  \small
\caption{The accuracy (\%) of ResNet18 on ImageNet-1k with various teachers. Results of other KD methods refer to the papers of CRD~\cite{crd} and ESKD~\cite{cho2019efficacy}. ATKD $_{A_{R34}}$~\cite{ATKD} denotes ResNet34 used as the assistant teacher. N/A means no available results. Our DisWOT
obtains better performance than other methods and improves students' performance positively correlated with that of the teacher.}
  \resizebox{1\linewidth}{!}{
    \begin{tabular}{ll|l|l|lllllll|l}
    \toprule
    Teacher & Student & Acc.   & Teacher & Student & KD~\cite{ref10_kd}    & ESKD~\cite{cho2019efficacy}  & $ATKD_{A_{R18}}$~\cite{ATKD}  & ONE~\cite{ref17_one}   & DML~\cite{ref15_dml}   & CRD~\cite{crd}  & DisWOT \\
    \midrule
    \multirow{2}[2]{*}{ResNet34} & \multirow{2}[2]{*}{ResNet18} & Top-1 & 73.40 & 69.75 & 70.66 & 70.89 & 70.78 & 70.55 & 71.03 & 71.17 & \textbf{72.08} \\
          &       & Top-5 & 91.42 & 89.07 & 89.88 & 90.06 & 89.99 & 89.59 & 90.28 & 90.32 & \textbf{90.38}   \\
    \midrule
    Teacher & Student & Acc.   & Teacher & Student & KD~\cite{ref10_kd}    & $ATKD_{A_{R18}}$~\cite{ATKD}& $ATKD_{A_{R34}}$~\cite{ATKD}   & Seq. ESKD~\cite{cho2019efficacy} & ESKD~\cite{cho2019efficacy}   &  SRRL~\cite{yang2021knowledge}   & DisWOT  \\
    \midrule
    \multirow{2}[2]{*}{ResNet50} & \multirow{2}[2]{*}{ResNet18} & Top-1 & 76.16 & 69.75 & 70.68 & 70.65 & 70.85 & 70.65 & 70.95 & 71.20 &  \textbf{72.30}      \\
          &       & Top-5 & 92.86 & 89.07  &  N/A & N/A & N/A & N/A & N/A & N/A &   \textbf{90.51}     \\
    \bottomrule
    \end{tabular}
    }
  \label{tab:imagenet}%
\end{table*}%

\subsection{Experiments on NAS-Bench-201}\label{exp:search on 201}

\noindent\textbf{Implementation Details.} For search trials, we first adopt ResNet110/56 as the teachers and then Conduct an evolution search with the DisWOT metric and get the best student network. We randomly sampled 50 candidate architectures to evaluate sequencing consistency. The distillation settings are the same as the Sec.\ref{exp:s0}

\noindent\textbf{Comparison results} As shown in Table~\ref{tab:nb201_sota}, some training-free methods can achieve good results with much faster speedups, such as NWOT and TE-NAS. Our proposed method DisWOT achieves a speedup ratio of 180$\times$, where if semantic similarity metric is removed, we can achieve a 300$\times$ speedup ratio at the expense of some accuracy.

\subsection{Experiments on ImageNet}\label{exp:ImageNet}

\noindent\textbf{Implementation Details}. We searched the ResNet18 level network regarding the search space in NDS~\cite{Radosavovic2020DesigningND}. Specifically, we limit the number of parameters to less than 13M and the depth of the network to up to 20 layers and find the optimal network by evolution algorithm with the DisWOT metric. As shown in Table~\ref{tab:imagenet}, guided by three different sizes of networks, we used DisWOT to find the optimal student network. We trained the student network obtained by the search using the distillation strategy in DisWOT. Implementation details are available in supplementary materials. 

\begin{table}[htbp]
  \centering
\caption{Spearman correlation ( ''mean$\pm$std'') of DisWOT on search space $S_0$.}
    \begin{tabular}{lll}
    \toprule
    Knowledge  & Metric & Spearman (\%)\\
    \midrule
    $\mathcal{M}_{s}$  &   FitNets~\cite{ref11_feature_kd}    & 64.06$\pm$6.11 \\
    $\mathcal{M}_{s}$  &   Similarity matrix   & 73.68$\pm$5.45 \\
    $\mathcal{M}_{r}$ &   RKD ~\cite{ref14_relation_crd}     & 13.52$\pm$11.51 \\
    $\mathcal{M}_{r}$ &    Similarity matrix   & 72.36$\pm$3.42 \\
    $\mathcal{M}_{s}$ \& $\mathcal{M}_{r}$ &   Similarity matrix    & \textbf{77.51$\pm$2.76} \\
    \bottomrule
    \end{tabular}%
  \label{tab:Ablation}%
\end{table}%

\noindent\textbf{Comparison Results}. Table~\ref{tab:imagenet} reports the performance of DisWOT on ImageNet with ResNet34/50 as teacher network. The results demonstrate that the student architecture of the ResNet18-level obtained by DisWOT under different teacher guidance and using different distillation strategies yielded significantly better results than its counterparts.

\subsection{Ablation Studies of DisWOT}\label{exp:ablation}

We perform ablation experiments to verify the validity of each component of DisWOT in search space $S_0$. As shown in Table~\ref{tab:Ablation}, for semantic knowledge,  similarity matrix obtains a more robust ranking improvement than simple FitNet~\cite{ref11_feature_kd}. For $\mathcal{M}_{r}$, similarity matrix performs better on relational knowledge than RKD~\cite{ref13_rkd}. DisWOT integrates semantic and relational knowledge to obtain an additional ranking improvement than stand-alone scores. The weight initialization scheme plays an important role in zero-proxy. We verify the effect of the initialization strategy of the network on the ranking consistency. The results in Tab.~\ref{tab:initialization2} demonstrate that the Gaussian initialization strategy is detrimental to $\mathcal{M}_{s}$, but beneficial to $\mathcal{M}_{r}$.

\section{Conclusion}
In this paper, we present DisWOT, a new teacher-aware 
student architecture search without training framework for distillation. Based on key observations about the difference between vanilla and distillation accuracy, DisWOT measures the new zero-cost proxy conditioned on the similarity of feature semantics and sample relations between random-initialized teacher-student network. Then, DisWOT search

\begin{table}[htbp]
\small
\centering
\caption{``mean$\pm$std \%'' Spearman of proxies via Kaiming and Gaussian initialization on search space $S_0$ and  NAS-Bench-201  with various seeds.}
\resizebox{1\linewidth}{!}{
\begin{tabular}{llllll}
\hline
Space                    & Initial   & Fisher      & GradNorm       & NWOT       & DisWOT       \\ \hline
\multirow{2}{*}{$S_0$}      & Kaim.  & 81.37$\pm$0.01  & 82.35$\pm$0.01 & 45.66$\pm$0.05 & 84.08$\pm$0.03 \\
                         & Gauss. & 80.99$\pm$0.01  & 75.50$\pm$0.01 & 45.36$\pm$0.03 & 91.38$\pm$0.03   \\ 
\multirow{2}{*}{NB-201} & Kaim.  & 54.63$\pm$0.15  & 58.70$\pm$0.11 & 64.41$\pm$0.08 & 65.57$\pm$0.02   \\
                         & Gauss. & 45.91$\pm$0.09  & 45.70$\pm$0.11 & 62.24$\pm$0.07 & 72.36$\pm$0.02   \\ \hline
\end{tabular}
}
\label{tab:initialization2}
\end{table}

\noindent for the best student architectures for the given teacher using an evolutionary algorithm with these metrics. 
Thorough evaluations are performed on diverse datasets and search spaces, and DisWOT achieves significant performance gains in various neural networks with at least 180$\times$ training acceleration. We experimentally and theoretically explained the relationship between similarity difference and distillation performance. In addition, we also extend DisWOT to new distillers and general zero proxy to predict the performance of models. By doing this, we bridge the higher-order knowledge bewteen distillation and network architecture search.  This approach represents an elegant and practical solution, which we hope will inspire future research on knowledge distillation and architecture search design.

\noindent\textbf{Limitations.} Following most zero-cost NAS, we  evaluate DisWOT in classification tasks. In the future work, we will make efforts to expand the DisWOT for downstream tasks~(e.g., object detection and semantic segmentation).

\appendix

\section{More Comparisons and Discussions}

In this section, we provide more analysis and discussion about DisWOT from different aspects.

\subsection{DisWOT under different teacher models.}
When the teacher model becomes larger, the fixed hand-designed model would have huge teacher-student gaps, limiting the performance gain. DisWOT aims to solve this problem by searching the suitable student architecture  for different teacher models. According to the results in Table~\ref{tab:different_teacher}, the accuracy of the student network ResNet20~\cite{resnet} is unable to make consistent gains as the size of the teacher network increases. Our proposed DisWOT enables a consistent increase in student network performance as the teacher network capacity increases on search space $S_0$. In addition, DisWOT$\dag$ achieves a performance gain of about 2\% when stronger distillers are adopted.

\begin{table}[htbp]
	\centering
	\caption{Top-1 accuracy (\%) of ResNet20 with KD~\cite{kd}, student  (DisWOT) with KD~\cite{kd}, student  (DisWOT) with DisWOT$\dag$ on search space $S_0$ under different teachers. }
	\begin{tabular}{lccc}
		\toprule
		Teacher   & ResNet20 & DisWOT & DisWOT$\dag$ \\
		\midrule
		ResNet32  & 70.24    & 71.01  & 71.85        \\
		ResNet44  & 70.56    & 71.25  & 72.12        \\
		ResNet56  & 70.98    & 71.63  & 72.56        \\
		ResNet110 & 70.79    & 71.84  & 72.92        \\
		\bottomrule
	\end{tabular}%
	\label{tab:different_teacher}%
\end{table}%

\begin{table*}[htbp]
	\centering
	\caption{Details experiments of the discrepancy between vanilla accuracy and distillation accuracy on search space $S_0$.}
	\resizebox{1\linewidth}{!}{
		\begin{tabular}{llll|lll|lll}
			\toprule
			\multirow{2}[4]{*}{Method} & \multicolumn{3}{c|}{Ranking with distill accuracy} & \multicolumn{3}{c|}{Ranking with vanilla accuracy} & \multicolumn{3}{c}{Ranking gap for distill and vanilla accuracy}                                                                                                                             \\
			\cmidrule{2-10}            & Kendall's Tau                                      & Spearman                                           & Pearson                                                          & Kendall's Tau & Spearman & Pearson & Kendall's Tau              & Spearman                   & Pearson                    \\
			\midrule
			FLOPs~\cite{zeronas}       & 51.61                                              & 72.92                                              & 76.40                                                            & 58.74         & 79.47    & 79.19   & 7.13~(\red{$\downarrow$})  & 6.55~(\red{$\downarrow$})  & 2.79~(\red{$\downarrow$})  \\
			Fisher~\cite{zeronas}      & 62.86                                              & 81.37                                              & 20.90                                                            & 81.68         & 95.28    & 70.24   & 18.82~(\red{$\downarrow$}) & 13.91~(\red{$\downarrow$}) & 49.34~(\red{$\downarrow$}) \\
			Grad\_norm~\cite{zeronas}  & 63.75                                              & 82.35                                              & 39.35                                                            & 84.76         & 96.55    & 76.07   & 21.01~(\red{$\downarrow$}) & 14.2~(\red{$\downarrow$})  & 36.72~(\red{$\downarrow$}) \\
			NWOT~\cite{NWOT}           & 31.87                                              & 45.66                                              & 48.99                                                            & 40.29         & 56.46    & 56.23   & 8.42~(\red{$\downarrow$})  & 10.80~(\red{$\downarrow$}) & 7.24~(\red{$\downarrow$})  \\
			Plain~\cite{zeronas}       & 10.72                                              & 13.57                                              & -0.91                                                            & 54.98         & 77.12    & 67.55   & 44.26~(\red{$\downarrow$}) & 63.55~(\red{$\downarrow$}) & 68.46~(\red{$\downarrow$}) \\
			SNIP~\cite{snip}           & 67.22                                              & 85.07                                              & 51.09                                                            & 84.66         & 96.38    & 77.83   & 17.44~(\red{$\downarrow$}) & 11.31~(\red{$\downarrow$}) & 26.74~(\red{$\downarrow$}) \\
			\textbf{DisWOT}            & 73.98                                              & 91.38                                              & 84.83                                                            & 73.02         & 91.26    & 82.98   & 0.96~(\blue{$\uparrow$})   & 0.12~(\blue{$\uparrow$})   & 1.85~(\blue{$\uparrow$})   \\
			\bottomrule
		\end{tabular}%
	}\label{tab:kd_of_zc_proxies}
\end{table*}

\subsection{Ranking correlation metrics}

We denote the ground-truth (GT) performance and approximated scores of architectures $\alpha_i(i=1,...,N)$ as $\beta_i(i=1,...,N)$ and $\gamma_i(i=1,...,N)$, respectively, and the ranking of the GT and estimated score $\beta_i$, $\gamma_i$ as $r_i,k_i\in \{1,...,N\}$. Three correlation criteria is adopted in this paper. Pearson coefficient ($r$),   Kendall's Tau ($\tau$),  Spearman coefficient ($\rho$).

\begin{itemize}
	\item Pearson correlation coefficient (Linear Correlation):
	      $r=\mbox{corr}(\beta, \gamma)/\sqrt{\mbox{corr}(\beta, \beta) \mbox{corr}(\gamma, \gamma)}$.
	\item Kendall's Tau correlation coefficient: The relative difference of concordant pairs and discordant pairs $\tau=\sum_{i < j} \text{sgn}(\beta_i - \beta_j)\text{sgn}(\gamma_i - \gamma_j) / \binom{M}{2} $. 
	\item Spearman correlation coefficient: The Pearson correlation coefficient between the ranking variables $\rho=\mbox{corr}(r, k) / \sqrt{\mbox{corr}(r, r)\mbox{corr}(k, k)}$.
\end{itemize}

Pearson measures the linear relationship between two variables, while Kendall's Tau and Spearman measure the monotonic relationship. They return a value between -1 and 1, with -1 indicating an inverse correlation, 1 indicating a positive correlation, and 0 representing no relationship.

In search space $S_0$, we evaluate and verify the ranking consistency for all the architectures in the search space. In search space $S_1$, we randomly sampled 50 sub-networks from the search space to calculate the ranking consistency results due to the excessive time overhead of the full measurement and repeated each experiment 10 times.

\subsection{Detailed analysis about vanilla-distilling training disparity}
In this paper, we notice an interesting and non-trivial observation: the discrepancy between the model's performance under vanilla training and under distillation training. We present more analysis in detail here from three aspects.

\noindent\textbf{Ranking correlation degrade.} In Table~\ref{tab:kd_of_zc_proxies}, we tabulate the ranking correlations for different zero-proxies with distilling and vanilla training on the search space $S_0$. The existing zero-proxies' distillation correlations are reduced by $5\% \sim 69\%$ than the vanilla correlation. This common issue shows that existing zero-shot NAS methods score sub-optimal student architectures for a given teacher model. DisWOT not only has a better correlation for distillation but also is free of vanilla-distill ranking consistency degradation.
\begin{figure}
	\setlength{\abovecaptionskip}{0.cm}
	\setlength{\belowcaptionskip}{-0.cm}
	\centering
	\begin{minipage}[t]{0.50\linewidth}
		\centering
		\includegraphics[width=1\linewidth]{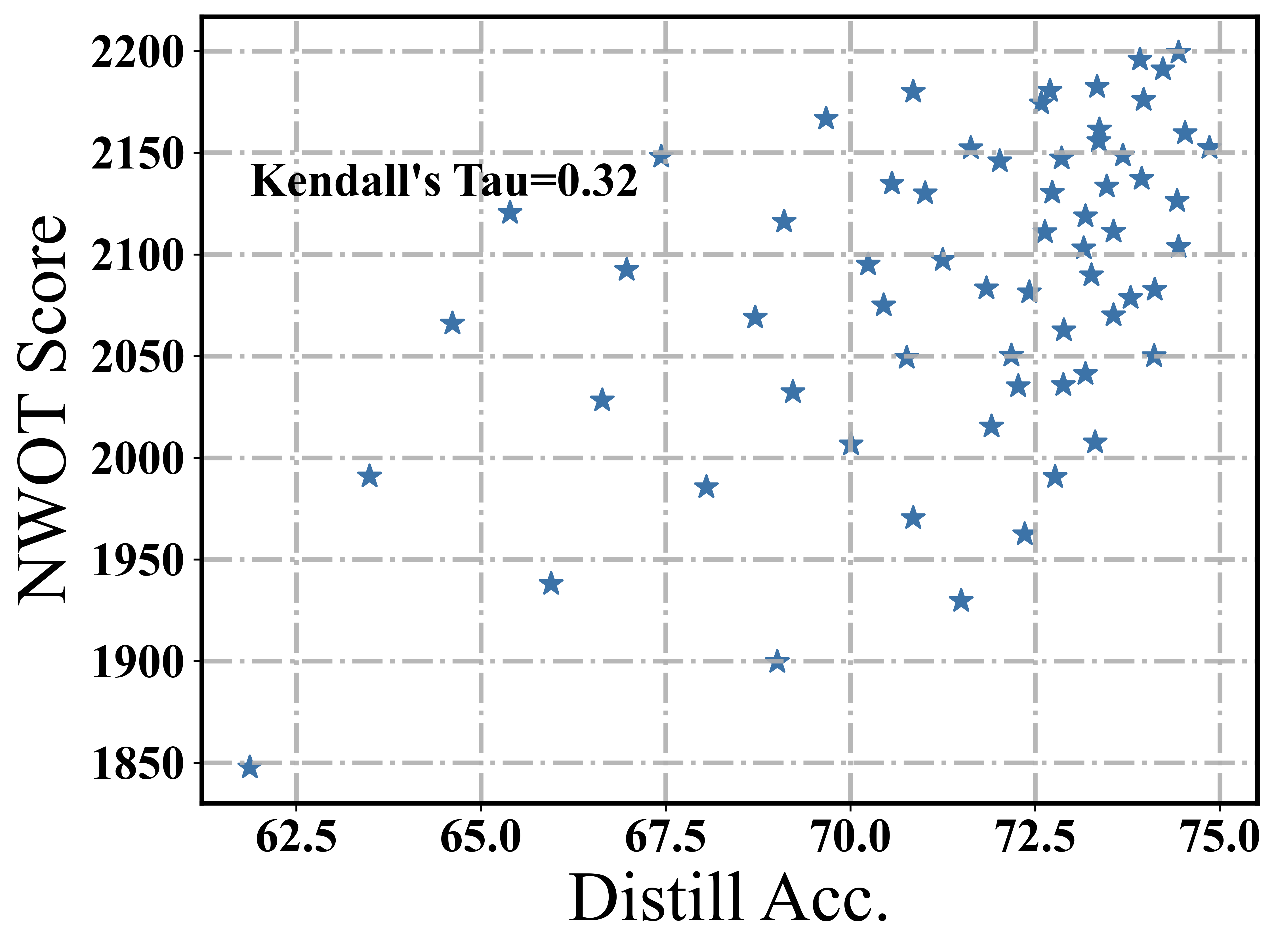}
	\end{minipage}%
	\begin{minipage}[t]{0.50\linewidth}
		\centering
		\centering
		\includegraphics[width=1\linewidth]{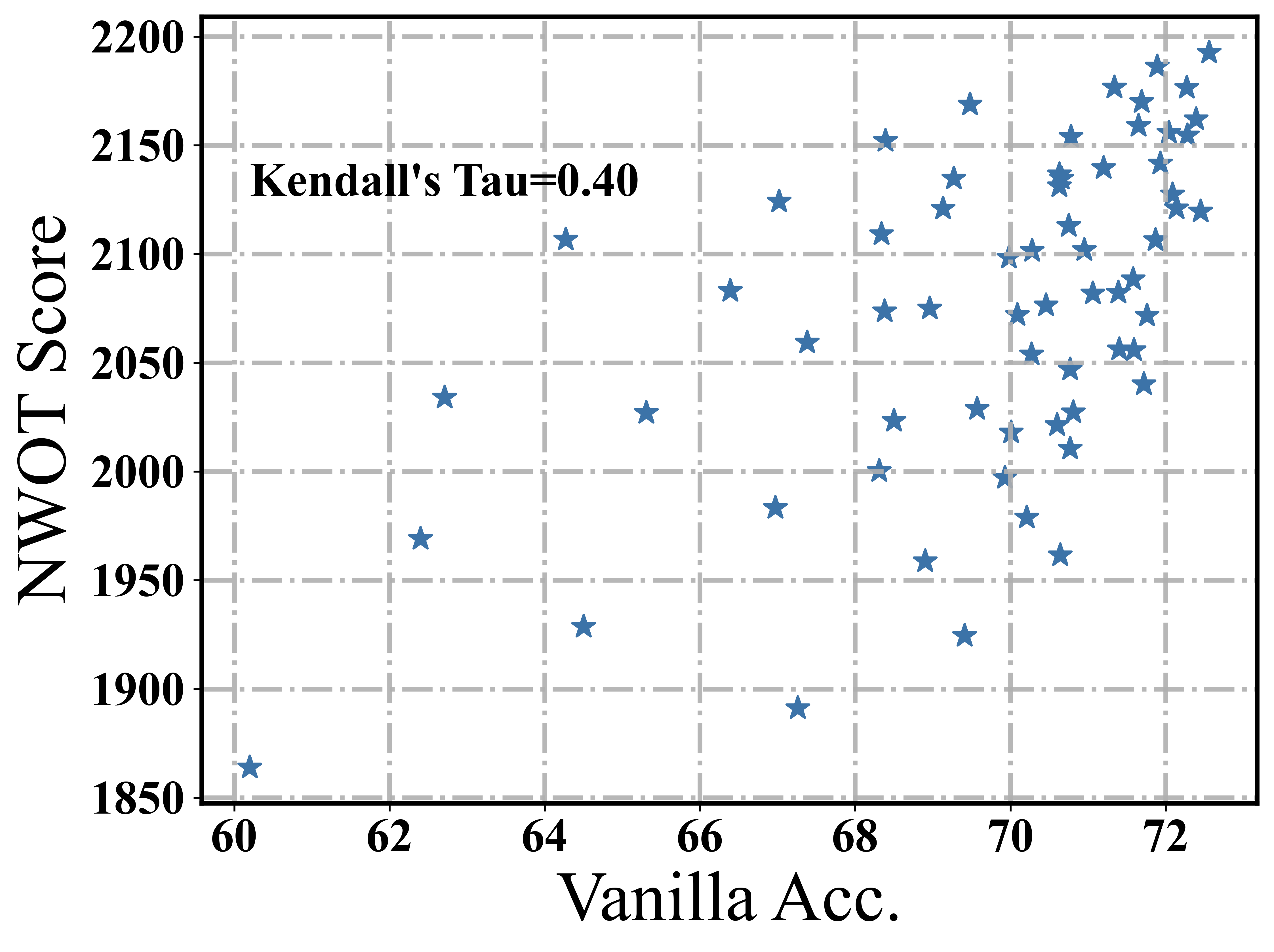}
	\end{minipage}
	\vspace{-0.1cm}
	\caption{Left: Correlation of distill accuracy \& NWOT score on search space $S_0$. Right: Correlation of vanilla accuracy \& NWOT score on search space $S_0$.}
	\label{fig:nwotx2}
\end{figure}

\begin{figure}
	\setlength{\abovecaptionskip}{0.cm}
	\setlength{\belowcaptionskip}{-0.cm}
	\centering
	\begin{minipage}[t]{0.47\linewidth}
		\centering
		\includegraphics[width=1\linewidth]{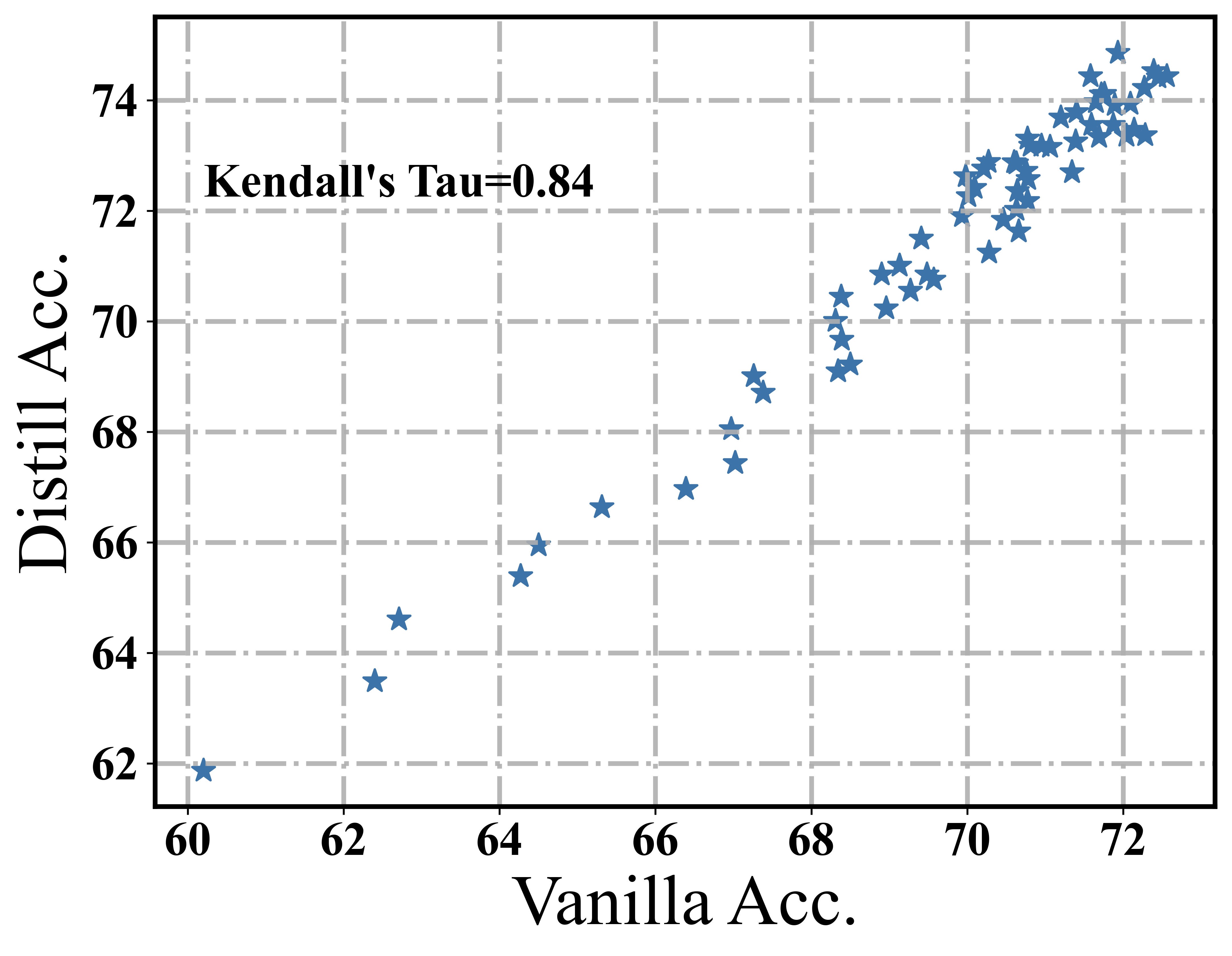}
	\end{minipage}%
	\begin{minipage}[t]{0.53\linewidth}
		\centering
		\centering
		\includegraphics[width=1\linewidth]{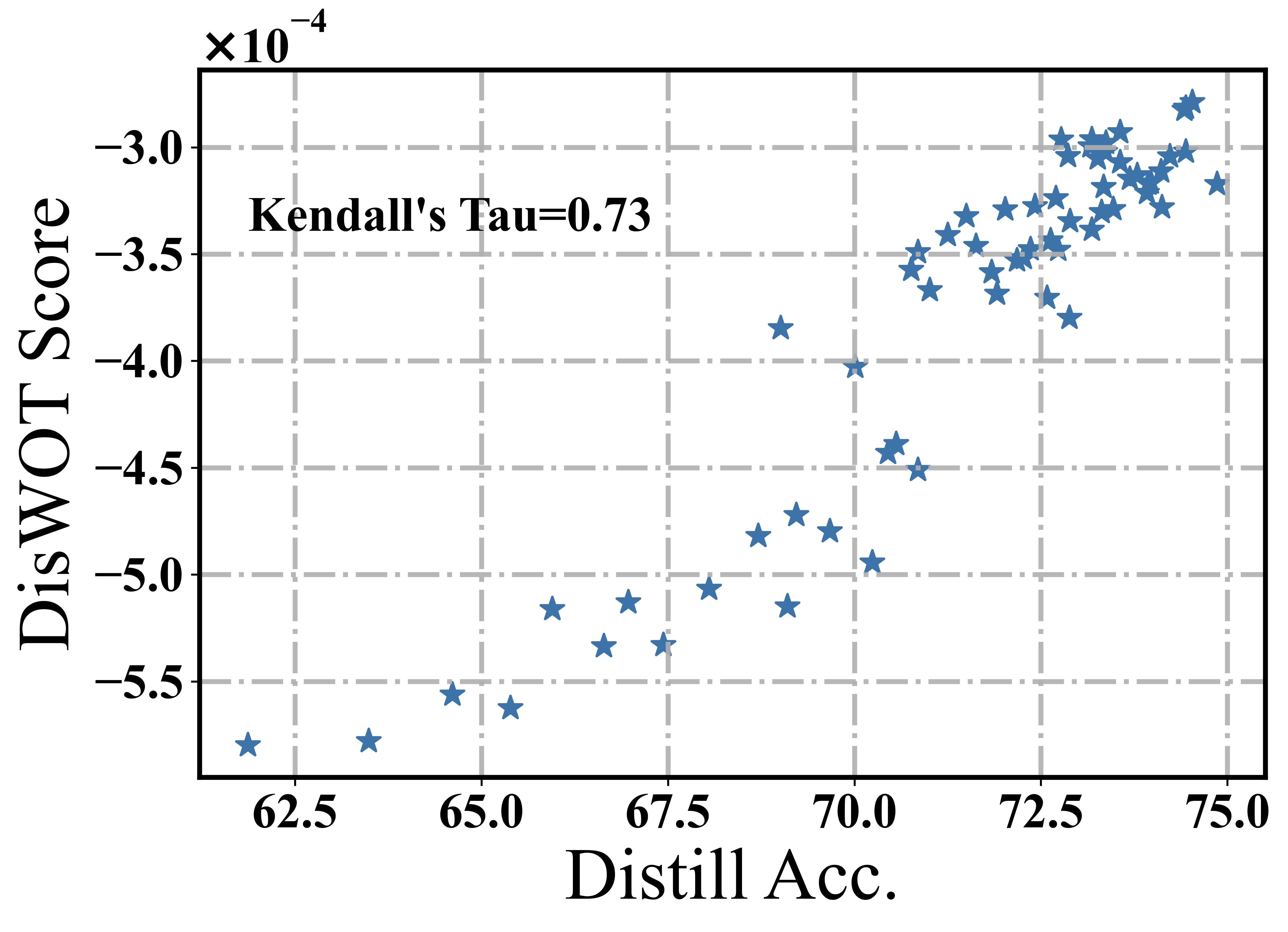}
	\end{minipage}
	\vspace{-0.1cm}
	\caption{Left: Correlation of vanilla accuracy \& distill accuracy on search space $S_0$. Right: Correlation of distilling accuracy \& DisWOT score on search space $S_0$.}
	\label{fig:discrepency}
\end{figure}

\noindent\textbf{Correlation visualization of different scores.} Figure~\ref{fig:nwotx2} demonstrates the ranking consistency of NWOT~\cite{NWOT} for distillation accuracy and vanilla accuracy, and there is a large discrepancy between them, with an 8\% difference in Kendall's Tau. Figure ~\ref{fig:discrepency} (left) demonstrates the ranking correlation between vanilla accuracy and distillation accuracy. Surprisingly, their correlation is only 0.84, which indicates that there is a non-negligible gap between distillation results and vanilla results. On the one hand, it indicates that a new zero-cost metric needs to be designed to improve the ranking consistency for the distillation. On the other hand, vanilla accuracy can be used as a rough measure of ranking consistency when distillation accuracy is unavailable. Figure~\ref{fig:discrepency} (right) illustrates that our proposed DisWOT achieve better ranking consistency of distillation accuracy on search space $S_0$.

\noindent\textbf{Analysis of detailed examples.} Table~\ref{tab:examples} summarizes 4 groups of specific student pairs with vanilla-distill gaps. For groups A and B, despite student models A2, B2 having more parameters and better vanilla accuracy, their distillation accuracy is inferior to A1, B1. This indicates an important effect on the overall depth of students for distillation. For groups C and D, the results show that the model with more blocks in stage-2 enjoys better distillation accuracy. In addition, DisWOT predicts the correct scores for these models.

\begin{table}[htbp]
	\centering
	\caption{ Parameters (K), vanilla accuracy (\%), distillation accuracy (\%), and prediction scores ($10^{-4}$)  of DisWOT for the student on search space $S_0$.}
	\resizebox{0.95\linewidth}{!}{
		\begin{tabular}{llcccc}
			\toprule
			Group & Student       & Param. & Vanilla Acc. & Distill Acc. & DisWOT \\
			\midrule
			A1    & ResNet[7,1,3] & 259.89 & 69.13        & 71.01        & 4.41   \\ 
			A2    & ResNet[3,3,3] & 278.32 & 69.57        & 70.76        & 3.34   \\ 
			\midrule
			B1    & ResNet[7,5,3] & 334.13 & 70.76        & 72.58        & 5.44   \\ 
			B2    & ResNet[1,7,3] & 343.22 & 70.77        & 72.18        & 5.20   \\ 
			\midrule
			C1    & ResNet[5,5,7] & 620.72 & 71.93        & 74.86        & 7.37   \\ 
			C2    & ResNet[3,7,7] & 648.50 & 72.45        & 74.42        & 7.33   \\ 
			\midrule
			D1    & ResNet[7,3,5] & 444.98 & 72.04        & 73.36        & 4.85   \\ 
			D2    & ResNet[5,5,5] & 472.76 & 72.09        & 73.94        & 8.17   \\ 
			\bottomrule
		\end{tabular}%
	}
	\label{tab:examples}%
\end{table}%

\subsection{Semantic properties of random networks}
\begin{figure}
	\centering
	\includegraphics[width=0.95\linewidth]{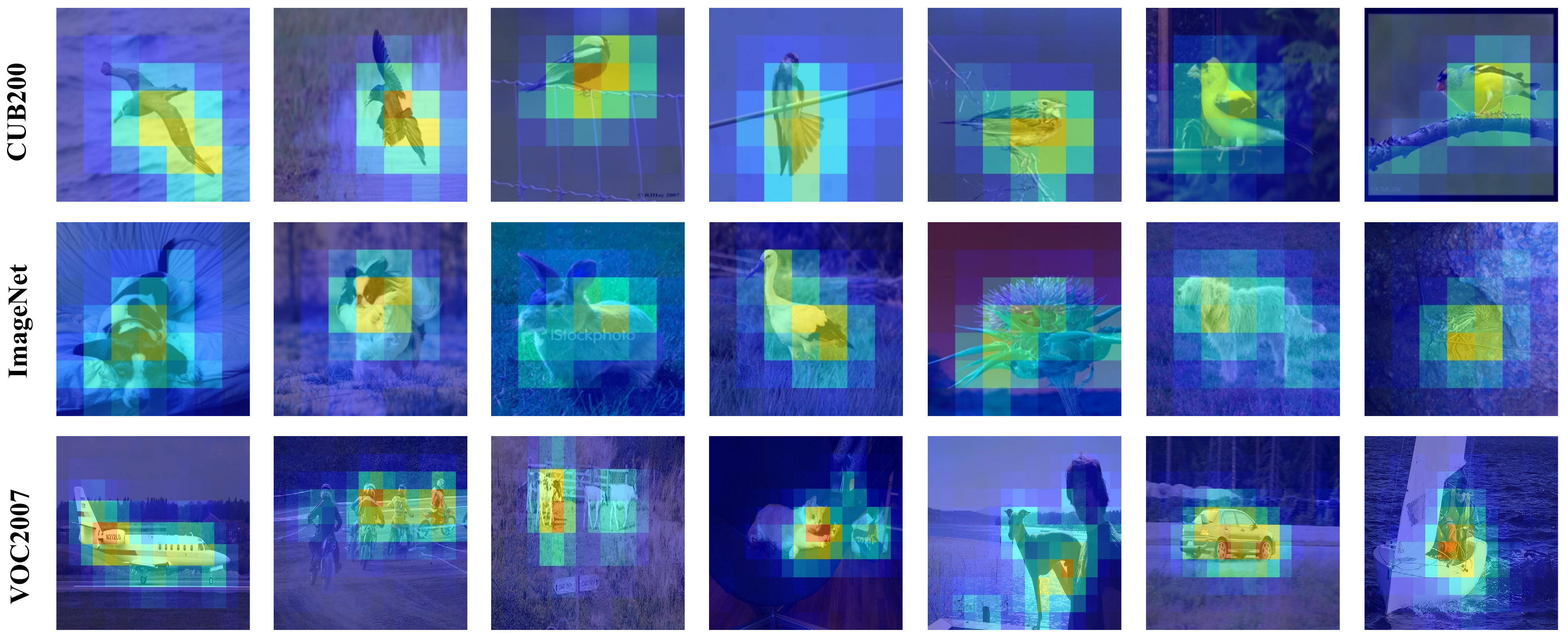}
	\caption{Localization results~\cite{random} of a randomly initialized network on ImageNet, VOC2007, and CUB-200. The network can localize the objects in an image, with a small standard deviation between different trials. Note that this figure is from Tobias~\cite{random}.}
	\label{fig:random_cnn}
\end{figure}

DisWOT leverages the semantic similarity metric of a randomly initialized teacher-student model to predict distillation performance, abandoning the training-based NAS paradigm. Models with various architectures have different semantic features because of their different effective receptive fields. To represent semantic and localization information, Gradient-weighted Class Activation Mapping (Grad-CAM~\cite{Grad}) methods have been widely adopted in weakly supervised object localization and model interpretability \cite{Bae2020RethinkingCA}.  Recently, several studies~\cite{scda,Bae2020RethinkingCA} reveal that randomly initialized models also have favorable semantic localization capabilities.  As shown in the visual localization heatmaps of Figure~\ref{fig:random_cnn}, we can intuitively observe that randomly initialized networks can locate a single object without any training. The visualization results demonstrate that semantic information exists even in random networks, which can localize the objects in an image. In addition, we evaluate different strategies~(e.g. CAM, Grad-CAM, and SCDA~\cite{scda}) for semantic localization maps and find that Grad-CAM achieves stable prediction performance for the semantic similarity metric. Thus, we adopt the Grad-CAM of deeper layers to capture the informative relation similarity in this paper.

\subsection{Comparisons with different NAS approaches.}

We presents the results of more NAS methods in this section and find that our proposed DisWOT is better able to distinguish good architectures than both One-shot NAS and Zero-shot NAS. As shown in Table~\ref{tab:vanilla_nas} and ~\ref{tab:distill_nas}, we compare DisWOT with one-shot NAS (e.g. ENAS~\cite{ENAS}, SETN~\cite{SETN}, SPOS~\cite{guo2019single}) and zero-shot NAS(a.k.a. Zen-NAS~\cite{ZenNAS}). Results demonstrate that DisWOT achieves better performance than its counterparts in distillation and classification results. In addition, we conduct more correlation evaluation in Tab.~\ref{tab:ranking}. The results show that DisWOT surpasses other proxies in the NAS-Bench-101/101-KD/201-KD and DisWOT ($\mathcal{M}_{s}$) achieves superior correlations than DisWOT ($\mathcal{M}_{r}$), which are consistent with findings in space $S_0$.

\begin{table}[htbp]
	\centering
	\small
	\caption{Top-1 accuracy (\%) of different NAS algorithms under distill training on NAS-Bench-201~\cite{nas201}.}
	\resizebox{1\linewidth}{!}{
		\begin{tabular}{llllll}
			\toprule
			Datasets       & ENAS~\cite{ENAS} & SETN~\cite{SETN} & SPOS~\cite{guo2019single} & Zen-NAS~\cite{ZenNAS} & DisWOT         \\
			\midrule
			CIFAR-10       & 34.94            & 81.61            & 92.98                     & 89.60                 & \textbf{93.55} \\
			CIFAR-100      & 11.14            & 59.78            & 72.91                     & 71.86                 & \textbf{74.21} \\
			ImageNet16-120 & 11.53            & 29.91            & 43.50                     & 39.44                 & \textbf{47.30} \\
			\bottomrule
		\end{tabular}%
	}
	\label{tab:distill_nas}%

\end{table}%

\begin{table}[t]
    \centering
    \small
    \caption{``mean$\pm$std \%'' Spearman correlation on NAS-Bench-101 and NAS-Bench-201. NAS-Bench-101/201-KD denotes to distill accuracies of the architectures on NAS-Bench-101/201.}
    \resizebox{1\linewidth}{!}{
        \begin{tabular}{lrrrr}
            \hline
            Method                     &  NAS-Bench-101          &  NAS-Bench-201         & NAS101-KD & NAS201-KD \\ \hline
            FLOPs                      & 30.81\%$\pm$0.00  & 63.38\%$\pm$0.06  & 15.56\%$\pm$0.04     & 64.55\%$\pm$0.01  \\
            Fisher                     & -38.81\%$\pm$0.14 & 35.91\%$\pm$0.09  & -33.92\%$\pm$0.14    & 4.45\%$\pm$0.08    \\
            Grad\_Norm                 & -39.23\%$\pm$0.08 & 58.70\%$\pm$0.11  & -39.16\%$\pm$0.01    & -10.01\%$\pm$0.11   \\
            SNIP                       & -29.01\%$\pm$0.09 & 58.17\%$\pm$0.15  & -21.78\%$\pm$0.02    & 16.91\%$\pm$0.10    \\
            Synflow                    & 43.69\%$\pm$0.12  & \textbf{74.61\%$\pm$0.08}  & 20.36\%$\pm$0.08     & 74.63\%$\pm$0.02    \\
            NWOT                       & 32.84\%$\pm$0.51  & 64.41\%$\pm$0.08  & 22.97\%$\pm$0.04     & 35.27\%$\pm$0.03    \\
            DisWOT ($\mathcal{M}_{s}$) & \textbf{49.61\%$\pm$0.05}  & 65.74\%$\pm$0.07  & 50.16\%$\pm$0.09     & 53.88\%$\pm$0.06    \\
            DisWOT ($\mathcal{M}_{r}$) & 30.74\%$\pm$0.06  & 56.46\%$\pm$0.08  & 42.94\%$\pm$0.11     & 45.27\%$\pm$0.07     \\
            \hline
        \end{tabular}
    }
\label{tab:ranking}
\end{table}
\subsection{About KD-based zero-cost proxies.}

In this section, we present DisWOT as a new universal zero proxy and propose a series of KD-based zero proxies based on this motivation. As shown in Table~\ref{tab:sp_kd_based_zc}, we further provide the ranking correlation of various KD-based zero-cost proxies on three datasets. We adopt the optimal architecture in the search space of NAS-Bench-201 as a teacher network and conduct 10 independent experiments of three knowledge distillation methods (a.k.a., CC~\cite{cc}, KD~\cite{kd}, and NST~\cite{nst}). We observe that DisWOT achieves an acceptable ranking correlation on three datasets. NST~\cite{nst} show impressive ranking ability, which is the best KD-based zero-proxies in DisWOT framework. DisWOT reveals better performance than most of zero-cost proxies under vanilla training, as shown in Table~\ref{tab:vanilla_nas}.

\begin{table}[htbp]
	\centering
	\caption{Ranking correlation of our KD-based zero-cost proxies on NAS-Bench-201.}
	\resizebox{0.9\linewidth}{!}{
		\begin{tabular}{llccc}
			\hline
			Datasets                    & Method         & \multicolumn{1}{l}{Kendall's Tau} & \multicolumn{1}{l}{Spearman} & \multicolumn{1}{l}{Pearson} \\ \hline
			\multirow{4}{*}{CIFAR-10}   & CC~\cite{cc}   & 0.48                              & 0.68                         & 0.56                        \\
			                            & KD~\cite{kd}   & 0.35                              & 0.50                         & 0.40                        \\
			                            & NST~\cite{nst} & 0.64                              & 0.83                         & 0.72                        \\
			                            & DisWOT         & 0.41                              & 0.61                         & 0.54                        \\ \hline
			\multirow{4}{*}{CIFAR-100}  & CC~\cite{cc}   & 0.43                              & 0.65                         & 0.58                        \\
			                            & KD~\cite{kd}   & 0.38                              & 0.54                         & 0.55                        \\
			                            & NST~\cite{nst} & 0.57                              & 0.72                         & 0.64                        \\
			                            & DisWOT         & 0.56                              & 0.72                         & 0.65                        \\ \hline
			\multirow{4}{*}{ImageNet16} & CC~\cite{cc}   & 0.53                              & 0.71                         & 0.66                        \\
			                            & KD~\cite{kd}   & 0.44                              & 0.61                         & 0.65                        \\
			                            & NST~\cite{nst} & 0.54                              & 0.74                         & 0.74                        \\
			                            & DisWOT         & 0.49                              & 0.69                         & 0.55                        \\ \hline
		\end{tabular}%
	}
	\label{tab:sp_kd_based_zc}%
\end{table}%

\begin{table}[htbp]
	\centering
	\caption{Top-1 accuracy (\%) of different NAS algorithms under vanilla training on NAS-Bench-201~\cite{nas201}.}
	\resizebox{1\linewidth}{!}{
		\begin{tabular}{llllll}
			\toprule
			Datasets       & ENAS~\cite{ENAS} & SETN~\cite{SETN} & SPOS~\cite{guo2019single} & Zen-NAS~\cite{ZenNAS} & DisWOT         \\
			\midrule
			CIFAR-10       & 53.89            & 87.64            & 93.23                     & 90.70                 & \textbf{93.37} \\
			CIFAR-100      & 13.96            & 59.05            & 71.03                     & 68.26                 & \textbf{71.53} \\
			ImageNet16-120 & 14.84            & 32.52            & 42.19                     & 40.60                 & \textbf{45.50} \\
			\bottomrule
		\end{tabular}%
	}
	\label{tab:vanilla_nas}%
\end{table}%

\section{Details of Search Space and Settings}

In this section, we introduce the implementation details in the three search spaces and the detailed training settings.

\subsection{$S_0$ search space}\label{s0}

\noindent\textbf{Search space.} As illustrated in Figure~\ref{tab:s0_search_space}, we construct the search space $S_0$ based on ResNet20, a simple resnet designed for CIFAR-10/100, where each building block consists of two $3\times 3$ convolutional layers and the depth of each residual block is searched in set \{1,3,5,7\}. The search space size is $4^3=64$ in total.

\noindent\textbf{Implementation details.} As for results of vanilla classification results, we train each architecture in the search space $S_0$ with the same strategy. For each architecture in the search space $S_0$, we adopt ResNet110 as a teacher network. Specifically, we train each architecture via momentum SGD, using cross-entropy loss for 240 epochs. We set the weight decay as 5e-4 and adopted a multi-stage scheduler to decay the learning rate from 0.1 to 0. We use the random flip with the probability of 0.5, the random crop $32\times 32$ patch with 4 pixels paddings, and the normalization over RGB channels. All of the experiments are based on CIFAR-100 datasets. As for the distillation results, the vanilla knowledge distillation methods~\cite{kd} are adopted. Specifically, we conduct experiments based on the CRD~\cite{crd}. For KD~\cite{kd}, we follow the Equation~\ref{eqa:kd} and set $\alpha=0.9$ and $\rho=4$.
\begin{equation}
	\mathcal{L}_{KL}=\alpha \rho^2 CE\left(\sigma\left(z^T / \rho\right), \sigma\left(z^S / \rho\right)\right)\label{eqa:kd}
\end{equation}
\noindent where $z^T$ and $z^S$ denote the logits of teacher and student, respectively. $\rho$ is the temperature, $\alpha$ is a balancing weight, and $\sigma$ is a softmax function. CE denotes the cross entropy loss.

\begin{table}[t]
	\caption{Supernet architecture of the $S_0$ search space. Each line describes a sequence of 1 or more identical layers, repeated \textit{repeat} times. All layers in the same sequence have the same number of output channels.}\label{supernet_arch}
	\centering
	\resizebox{1\linewidth}{!}{
		\begin{tabular}{c|c|c|c|c}
			\hline
			input          & block           & channels & repeat    & stride \\
			\hline
			$32^2\times3$  & $3\times3$ conv & 16       & 1         & 2      \\
			$32^2\times16$ & Res Block       & 16       & [1,3,5,7] & 2      \\
			$16^2\times16$ & Res Block       & 32       & [1,3,5,7] & 2      \\
			$8^2\times32$  & Res Block       & 64       & [1,3,5,7] & 2      \\
			$8^2\times64$  & Global Avgpool  & -        & 1         & -      \\
			64             & FC              & 100      & 1         & -      \\
			\hline
		\end{tabular}
	}
	\label{tab:s0_search_space}
\end{table}

\subsection{$S_1$ search space}

\noindent\textbf{Search space.} The search space $S_1$ is following the cell-based search space NAS-Bench-201~\cite{nas201}, where a cell is represented as a directed acyclic graph (DAG). Each edge in search space $S_1$ is associated with an operation selected from a predefined operation set, which consists of (1) zero, (2) skip connection, (3) $1\times 1$ convolution, (4) $3\times 3$ convolution, and (5) $3\times 3$ average pooling layer. The DAG has 4 nodes, each representing the sum of all features from previous nodes. The search space size of $S_1$ is 15,625 in total.

\noindent\textbf{Implementation Details.} We randomly sampled 50 candidate architectures to evaluate ranking consistency. All experiments are implemented on a single NVIDIA 3090Ti GPU, with the baseline from the AutoDL~\cite{nas201}. We recommend using a network with higher complexity or better performance in the search space as the teacher network. The process of DisWOT is divided into three steps: 
(1) Determine a specific teacher network (deeper or more complex).
(2) Perform an evolutionary search with DisWOT metrics to obtain the best student network.
(3) Distill the student network with vanilla KD~\cite{ref10_kd} based on a specific teacher network.
The distillation setting is the same as Section~\ref{s0}.

\noindent\textbf{Searched architectures.} For other NAS methods, the searched architectures (see Table~\ref{tab:nb201_results}) of RS~\cite{nas201}, ENAS~\cite{ENAS}, SETN~\cite{SETN} and SPOS~\cite{guo2019single} are borrowed from the official implementation~\cite{nas201}, and the remaining zero-shot NAS methods utilize the same evolutionary search algorithm. Expressly, we set the initial population size as 20, and the sample size as 10. The total evolution search cycle is set as 5,000. We calculate the zero-proxy score with only one batch of data as fitness during evolution. The teacher network used in DisWOT is the best architecture in the search space.

\begin{table*}[]
	\caption{Searched architectures of NAS algorithms. The searched results are denoted by a string from NAS-Bench-201 API~\cite{nas201}.}
	\small
	\begin{tabular}{l|l}
		\hline
		                          & \multicolumn{1}{c}{Searched architectures}                                                                                                         \\ \hline
		RS~\cite{nas201}          & |skip\_connect$\sim$0|+|nor\_conv\_3x3$\sim$0|skip\_connect$\sim$1|+|nor\_conv\_3x3$\sim$0|nor\_conv\_1x1$\sim$1|avg\_pool\_3x3$\sim$2|   \\
		ENAS~\cite{ENAS}          & |skip\_connect$\sim$0|+|avg\_pool\_3x3$\sim$0|skip\_connect$\sim$1|+|avg\_pool\_3x3$\sim$0|skip\_connect$\sim$1|skip\_connect$\sim$2|     \\
		SETN~\cite{SETN}          & |nor\_conv\_3x3$\sim$0|+|skip\_connect$\sim$0|skip\_connect$\sim$1|+|skip\_connect$\sim$0|skip\_connect$\sim$1|avg\_pool\_3x3$\sim$2|     \\
		SPOS~\cite{guo2019single} & |skip\_connect$\sim$0|+|nor\_conv\_1x1$\sim$0|nor\_conv\_3x3$\sim$1|+|nor\_conv\_1x1$\sim$0|avg\_pool\_3x3$\sim$1|nor\_conv\_3x3$\sim$2|  \\
		Zen-NAS~\cite{ZenNAS}     & |skip\_connect$\sim$0|+|nor\_conv\_3x3$\sim$0|nor\_conv\_3x3$\sim$1|+|skip\_connect$\sim$0|skip\_connect$\sim$1|nor\_conv\_3x3$\sim$2|    \\
		NWOT~\cite{NWOT}          & |nor\_conv\_1x1$\sim$0|+|nor\_conv\_3x3$\sim$0|nor\_conv\_1x1$\sim$1|+|nor\_conv\_1x1$\sim$0|nor\_conv\_3x3$\sim$1|nor\_conv\_1x1$\sim$2| \\
		DisWOT(ours)              & |skip\_connect$\sim$0|+|nor\_conv\_3x3$\sim$0|nor\_conv\_1x1$\sim$1|+|nor\_conv\_1x1$\sim$0|nor\_conv\_3x3$\sim$1|nor\_conv\_3x3$\sim$2|  \\ \hline
	\end{tabular}\label{tab:nb201_results}
\end{table*}

\subsection{$S_2$ search space}

\noindent\textbf{Search space.} Following NDS~\cite{Radosavovic2019OnND}, we design the search space for CIFAR and ImageNet, respectively. The search space designed for CIFAR consists of a stem, followed by 6 stages, and a head, as shown in Table~\ref{s2_cifar}. The $i$-th stage consists of $d_i$ blocks with $c_i$ channels and stride of $s_i\in \{1,2\}$. The number of channels $c_i$ needs to be divisible by 8, and the minimal number of channels should be larger than 8. The candidate blocks can be residual blocks or bottleneck blocks defined in ResNet, and the kernel size can be chosen from set \{3,5,7\}. As shown in Table~\ref{s2_imagenet}, the search space designed for ImageNet consists of 4 stages, following the configuration of ResNet18.

\begin{table}[H]
	\caption{Design space parameterization of $S_2$ for CIFAR-10/100. "POOL" denotes the global average pooling, and "FC" denotes a fully connected network.}
	\centering
	\resizebox{1\linewidth}{!}{
		\begin{tabular}{ccccc}
			\hline
			stage  & block            & \multicolumn{1}{l}{channels} & \multicolumn{1}{l}{repeat} & \multicolumn{1}{l}{stride} \\ \hline
			steam  & $3\times 3$ conv & $c_0$                        & 1                          & 1                          \\
			stage1 & \{block\}        & $c_1$                        & $d_1$                      & $s_1$                      \\
			stage2 & \{block\}        & $c_2$                        & $d_2$                      & $s_2$                      \\
			stage3 & \{block\}        & $c_3$                        & $d_3$                      & $s_3$                      \\
			stage4 & \{block\}        & $c_4$                        & $d_4$                      & $s_4$                      \\
			stage5 & \{block\}        & $c_5$                        & $d_5$                      & $s_5$                      \\
			stage6 & \{block\}        & $c_6$                        & $d_6$                      & $s_6$                      \\
			head   & POOL + FC        & 10                           & -                                                       \\ \hline
		\end{tabular}\label{s2_cifar}
	}
\end{table}

\begin{table}[H]
	\caption{Design space parameterization of $S_2$ for ImageNet. "POOL" denotes the global average pooling, and "FC" denotes fully connected network.}
	\centering
	\begin{tabular}{ccccc}
		\hline
		stage  & block            & \multicolumn{1}{l}{channels} & \multicolumn{1}{l}{repeat} & \multicolumn{1}{l}{stride} \\ \hline
		steam  & $7\times 7$ conv & $c_0$                        & 1                          & 2                          \\
		stage1 & \{block\}        & $c_1$                        & $d_1$                      & $s_1$                      \\
		stage2 & \{block\}        & $c_2$                        & $d_1$                      & $s_2$                      \\
		stage3 & \{block\}        & $c_3$                        & $d_1$                      & $s_3$                      \\
		stage4 & \{block\}        & $c_4$                        & $d_1$                      & $s_4$                      \\
		head   & POOL + FC        & 1000                         & -                          & -                          \\ \hline
	\end{tabular}\label{s2_imagenet}
\end{table}

\noindent\textbf{Training settings.} We search the ResNet18 level network regarding the search space in NDS~\cite{Radosavovic2019OnND}. Specifically, we limit the number of parameters to less than 13M and the depth of the network to up to 20 layers and find the optimal network by evolution algorithm with the DisWOT metric. Please refer to Section~\ref{sec:evolution_algo} for more details about the evolutionary search for the search space $S_2$. Specifically, we adopt ResNet34 as a teacher network and conduct a vanilla knowledge distillation process~\cite{ref10_kd} with $\rho=1$ and $\alpha=3$ as shown in Equation~\ref{eqa:kd}. For ImageNet, we follow the standard PyTorch practice, and the batch size is 256.

\noindent\textbf{Searched architectures.} After the evolutionary search, we presented the optimal student network obtained for CIFAR and ImageNet, as shown in the Table~\ref{tab:search_results_of_s2_cifar} and Table~\ref{tab:search_results_of_s2_img}, respectively. We observe that the searched architecture of DisWOT has very different characteristics from the artificially designed student architecture, i.e., DisWOT prefers student networks with larger convolutional kernels in shallow layers. Generally speaking, teacher networks tend to have deeper layers and thus have a larger receptive field. Guided by the teacher network, DisWOT favors larger convolutional kernels in the shallow layers so that the receptive field of the student network is as close to that of the teacher as possible. However, the network searched on ImageNet only changed the number of channels under the parameter restriction. We infer that expanding the kernel size leads to a massive amount of additional parameters, which will lead to exceeding the budget. We infer that the network will prefer a larger kernel if a sufficient budget is available.

\begin{table}[h]
	\centering
	\caption{Search results of the ResNet-like search space for CIFAR-10/100. "Basic" denotes the basic block proposed in ResNet~\cite{resnet}.}
	\label{supernet1}
	\resizebox{1\linewidth}{!}{
		\begin{tabular}{c|c|c|c|c} \hline
			{input}          & {block}           & {channels} & {repeat} & {stride} \\ \hline
			$32^2\times 3$   & $3\times 3$ conv  & 88         & 1        & 1        \\
			$32^2\times 88$  & Basic $7\times 7$ & 96         & 3        & 1        \\
			$32^2\times 120$ & Basic $5\times 5$ & 192        & 2        & 2        \\
			$16^2\times 192$ & Basic $5\times 5$ & 176        & 2        & 1        \\
			$16^2\times 96$  & Basic $5\times 5$ & 168        & 3        & 2        \\
			$8^2\times 168$  & Basic $3\times 3$ & 112        & 3        & 2        \\
			$4^2\times 112$  & Basic $3\times 3$ & 512        & 1        & 1        \\
			$512$            & POOL + FC         & 1000       & 1        & -        \\ \hline
		\end{tabular}\label{tab:search_results_of_s2_cifar}
	}
\end{table}

\begin{table}[h]
	\centering
	\caption{Search results of the ResNet-like search space for ImageNet under 13M parameter limit. "Basic" denotes the basic block proposed in ResNet~\cite{resnet}.}
	\label{supernet2}
	\resizebox{1\linewidth}{!}{
		\begin{tabular}{c|c|c|c|c} \hline
			{input}          & {block}           & {channels} & {repeat} & {stride} \\ \hline
			$224^2\times 3$  & $7\times 7$ conv  & 96         & 1        & 2        \\
			$112^2\times 96$ & Basic $3\times 3$ & 64         & 3        & 2        \\
			$56^2\times 64$  & Basic $3\times 3$ & 128        & 2        & 2        \\
			$28^2\times 128$ & Basic $3\times 3$ & 256        & 2        & 2        \\
			$14^2\times 256$ & Basic $3\times 3$ & 512        & 2        & 2        \\
			$512$            & POOL + FC         & 1000       & 1        & -        \\ \hline
		\end{tabular}\label{tab:search_results_of_s2_img}
	}
\end{table}

\begin{table}[!ht]
	\small
	\caption{DisWOT-11.7M based on ImageNet (Searched for segmentation task.)}
	\begin{center}
		\begin{tabular}{lllllll}
			\toprule
			block & kernel & in  & out  & stride & bottleneck & \# layers\tabularnewline
			\midrule
			\midrule
			Conv  & 7      & 3   & 64   & 2      & -          & 1\tabularnewline
			\midrule
			Res   & 3      & 64  & 64   & 2      & 64         & 2\tabularnewline
			\midrule
			Res   & 3      & 64  & 128  & 2      & 128        & 2\tabularnewline
			\midrule
			Res   & 3      & 128 & 256  & 2      & 256        & 2\tabularnewline
			\midrule
			Res   & 3      & 256 & 512  & 2      & 512        & 2\tabularnewline
			\midrule
			Conv  & 1      & 512 & 2384 & 1      & -          & 1\tabularnewline
			\bottomrule
		\end{tabular}
	\end{center}
	\label{tab:struct-ZenNet-0.1ms}
\end{table}

\section{Details of Algorithm for DisWOT}\label{implementation}

In this section, we describe the implementation details of the evolutionary alorhigtm and the implementation code of DisWOT.

\subsection{Implementation of Evolutionary Algorithm}\label{sec:evolution_algo}

Here we adopt Evolutionary Algorithm (EA) as an architecture generator to find optimal student network. In this section, we further describe the mutation process of the evolutionary algorithm in details. In the evolutionary algorithm, we randomly generate $P$ architectures with constraints $C$ and then select the $top-k$ architectures by DisWOT metric from the population. Then we randomly select the parent architecture from the $top-k$ architectures and mutate it. The mutation algorithm is presented in Algorithm ~\ref{mutation_algo}. Specifically, for $S_2$ search space, we provide basic blocks with a kernel size of \{3,5,7\} and choose \{0.67,0.8,1.25,1.5\} as the mutation range for channels. The number of channels should be divisible by 8, and the max number of channels is 2048. The depth of chosen block can be mutated in range \{+1, -1\}. After mutating the architecture, we check whether it is valid, e.g., the parameter is meed the predefined constraint.

\begin{algorithm}
	\small
	\caption{Mutation Algorithm for DisWOT}
	\textbf{Input}: Parent architecture $A_i$, Search space $S$.

	\textbf{Output}: Mutated architecture $\hat{P}_i$
	\begin{algorithmic}[1]
		\STATE Randomly select a block $a_i$ from Parent architecture $A_i$;
		\STATE Randomly mutate the kernel size of $a_i$ from $S(\{block\})$;
		\STATE Randomly mutate the width of $a_i$ from $S(c_i)$;
		\STATE Randomly mutate the depth of $a_i$ from $S(d_i)$;
		\STATE Check whether the mutated architecture is valid;
		\STATE Return the mutated architecture $\hat{P}_i$;
	\end{algorithmic}\label{mutation_algo}
\end{algorithm}

\subsection{Implementation of metric in DisWOT}

The section presents the implementation of semantic similarity metric and relation similarity metric in DisWOT. The semantic similarity metric measures the inter-correlation on the accumulated Grad-CAM for teacher and student networks, whose calculation needs one forward and one backward to get the localization information. The relation similarity metric measures the relationship between input samples whose activations of teacher and student networks are needed.

\noindent\textbf{Implementation of semantic similarity metric.} Here, we present the implementation code of the relation similarity metric, as shown in List~\ref{list:ickd}. We need the Grad-CAM maps of all classes for calculation, which needs at least one forward and one backward to get the Grad-CAM similarity. Different from ICKD~\cite{ickd}, there are mainly two differences: (1) The Gaussian initialized teacher network and student network is backpropagated only once. (2) We only use the grad of the fully-connected layer for calculation.

\begin{lstlisting}[language=Python, caption=The PyTorch implementation of semantic similarity metric.]
import torch
import torch.nn as nn 
import torch.nn.functional as F

def semantic_similarity_metric(teacher, student, batch_data):
    criterion = nn.CrossEntropyLoss() 
    image, label = batch_data 
    # Forward once.
    t_logits = teacher.forward(image)
    s_logits = student.forward(image)
    # Backward once.
    criterion(t_logits, label).backward()
    criterion(s_logits, label).backward()
    # Grad-cam of fc layer.
    t_grad_cam = teacher.fc.weight.grad
    s_grad_cam = student.fc.weight.grad
    # Compute channel-wise similarity
    return -1 * channel_similarity(t_grad_cam, s_grad_cam)

def channel_similarity(f_t, f_s):
    bsz, ch = f_s.shape[0], f_s.shape[1]
    # Reshape
    f_s = f_s.view(bsz, ch, -1)
    f_t = f_t.view(bsz, ch, -1)
    # Get channel-wise similarity matrix
    emd_s = torch.bmm(f_s, f_s.permute(0, 2, 1))
    emd_s = F.normalize(emd_s, dim=2)
    emd_t = torch.bmm(f_t, f_t.permute(0, 2, 1))
    emd_t = F.normalize(emd_t, dim=2)
    # Produce L_2 distance
    G_diff = emd_s - emd_t
    return (G_diff * G_diff).view(bsz, -1).sum() / (ch * bsz)

\end{lstlisting}\label{list:ickd}

\noindent\textbf{Implementation of relation similarity metric.} Here we present the implementation code of relation similarity, as shown in List~\ref{list:sp}. With only the logits of the teacher and student network, our relation similarity metric is easy to implement. There are mainly two differences compared with SP~\cite{sp}: (1) The teacher and student networks are initialized with kaiming initialization~\cite{kaimingInit}, which means the teacher network did not undergo any backpropagation. (2) Here, we only used the activation before global average pooling as input, and the activations of shallow layers are not utilized. In fact, we find that activations closer to the output are more informative. When using activations from shallow layers, experiments demonstrate that the relation similarity ranks poorly.

\begin{lstlisting}[language=Python, caption=The PyTorch implementation of relation similarity metric.]
import torch
import torch.nn as nn 
import torch.nn.functional as F

def relation_similarity_metric(teacher, student, batch_data):
    image, label = batch_data
    # Forward pass
    t_feats = teacher.forward_features(image)
    s_feats = student.forward_features(image)
    # Get activation before average pooling
    t_feat = t_feats[-2]
    s_feat = s_feats[-2]
    # Compute batch similarity
    return -1 * batch_similarity(t_feat, s_feat)

def batch_similarity(f_t, f_s):
    # Reshape
    f_s = f_s.view(f_s.shape[0], -1)
    f_t = f_t.view(f_t.shape[0], -1)
    # Get batch-wise similarity matrix
    G_s = torch.mm(f_s, torch.t(f_s))
    G_s = F.normalize(G_s)
    G_t = torch.mm(f_t, torch.t(f_t))
    G_t = F.normalize(G_t)
    # Produce L_2 distance
    G_diff = G_t - G_s
    return (G_diff * G_diff).view(-1, 1).sum() / (bsz * bsz)
\end{lstlisting}\label{list:sp}

{\small
\bibliographystyle{ieee_fullname}
\bibliography{egbib}

\begin{thebibliography}{10}\itemsep=-1pt

\bibitem{zeronas}
Mohamed~S Abdelfattah, Abhinav Mehrotra, {\L}ukasz Dudziak, and Nicholas~Donald
  Lane.
\newblock Zero-cost proxies for lightweight nas.
\newblock In {\em ICLR}, 2020.

\bibitem{ref43_vid}
Sungsoo Ahn, Shell~Xu Hu, Andreas Damianou, Neil~D Lawrence, and Zhenwen Dai.
\newblock Variational information distillation for knowledge transfer.
\newblock In {\em CVPR}, 2019.

\bibitem{Bae2020RethinkingCA}
Wonho Bae, Junhyug Noh, and Gunhee Kim.
\newblock Rethinking class activation mapping for weakly supervised object
  localization.
\newblock In {\em ECCV}, 2020.

\bibitem{baker2016designing}
Bowen Baker, Otkrist Gupta, Nikhil Naik, and Ramesh Raskar.
\newblock Designing neural network architectures using reinforcement learning.
\newblock In {\em ICLR}, 2017.

\bibitem{beyer2022knowledge}
Lucas Beyer, Xiaohua Zhai, Am{\'e}lie Royer, Larisa Markeeva, Rohan Anil, and
  Alexander Kolesnikov.
\newblock Knowledge distillation: A good teacher is patient and consistent.
\newblock In {\em CVPR}, 2022.

\bibitem{GPT-3}
Tom~B. Brown, Benjamin Mann, Nick Ryderand~Melanie Subbiah, Jared Kaplan,
  Prafulla Dhariwal, Arvind Neelakantan, Pranav Shyam, Girish Sastry, Amanda
  Askell, Sandhini Agarwal, Ariel Herbert-Voss, Gretchen Krueger, Tom Henighan,
  Rewon Child, Aditya Ramesh, Daniel M, Ziegler, Jeffrey Wu, Clemens Winter,
  Christopher Hesse, Mark Chen, Eric Sigler, Mateusz Litwin, Scott Gray,
  Benjamin Chess, Jack Clark, Christopher Berner, Sam McCandlish, Alec Radford,
  Ilya Sutskever, and Dario Amodei.
\newblock Language models are few-shot learners.
\newblock {\em arXiv preprint, arXiv:2005.14165}, 2020.

\bibitem{random}
Yun-Hao Cao and Jianxin Wu.
\newblock A random cnn sees objects: One inductive bias of cnn and its
  applications.
\newblock In {\em AAAI}, 2022.

\bibitem{chen2022knowledge}
Defang Chen, Jian-Ping Mei, Hailin Zhang, Can Wang, Yan Feng, and Chun Chen.
\newblock Knowledge distillation with the reused teacher classifier.
\newblock In {\em CVPR}, 2022.

\bibitem{lichengp}
Kunlong Chen, Liu Yang, Yitian Chen, Kunjin Chen, Yidan Xu, and Lujun Li.
\newblock Gp-nas-ensemble: a model for the nas performance prediction.
\newblock In {\em CVPRW}, 2022.

\bibitem{TENAS}
Wuyang Chen, Xinyu Gong, and Zhangyang Wang.
\newblock Neural architecture search on imagenet in four gpu hours: A
  theoretically inspired perspective.
\newblock In {\em ICLR}, 2020.

\bibitem{cho2019efficacy}
Jang~Hyun Cho and Bharath Hariharan.
\newblock On the efficacy of knowledge distillation.
\newblock In {\em ICCV}, 2019.

\bibitem{dong2023rd}
Peijie Dong, Xin Niu, Lujun Li, Zhiliang Tian, Xiaodong Wang, Zimian Wei,
  Hengyue Pan, and Dongsheng Li.
\newblock Rd-nas: Enhancing one-shot supernet ranking ability via ranking
  distillation from zero-cost proxies.
\newblock {\em arXiv preprint arXiv:2301.09850}, 2023.

\bibitem{linas2}
Peijie Dong, Xin Niu, Lujun Li, Linzhen Xie, Wenbin Zou, Tian Ye, Zimian Wei,
  and Hengyue Pan.
\newblock Prior-guided one-shot neural architecture search.
\newblock {\em arXiv preprint arXiv:2206.13329}, 2022.

\bibitem{nas201}
Xuanyi Dong and Yi Yang.
\newblock Nas-bench-201: Extending the scope of reproducible neural
  architecture search.
\newblock In {\em ICLR}, 2019.

\bibitem{SETN}
Xuanyi Dong and Yezhou Yang.
\newblock One-shot neural architecture search via self-evaluated template
  network.
\newblock {\em 2019 ICCV}, 2019.

\bibitem{dong2019search}
Xuanyi Dong and Yi Yang.
\newblock Searching for a robust neural architecture in four gpu hours.
\newblock In {\em CVPR}, 2019.

\bibitem{GDAS}
Xuanyi Dong and Yezhou Yang.
\newblock Searching for a robust neural architecture in four gpu hours.
\newblock {\em CVPR}, 2019.

\bibitem{VIT}
Alexey Dosovitskiy, Lucas Beyer, Alexander Kolesnikov, Dirk Weissenborn,
  Xiaohua Zhai, Thomas Unterthiner, Mostafa Dehghani, Matthias Minderer, Georg
  Heigold, Sylvain Gelly, et~al.
\newblock An image is worth 16x16 words: Transformers for image recognition at
  scale.
\newblock {\em arXiv preprint arXiv:2010.11929}, 2020.

\bibitem{Falkner2018BOHBRA}
Stefan Falkner, Aaron Klein, and Frank Hutter.
\newblock Bohb: Robust and efficient hyperparameter optimization at scale.
\newblock In {\em ICML}, 2018.

\bibitem{gu}
Jindong Gu and Volker Tresp.
\newblock Search for better students to learn distilled knowledge.
\newblock In {\em ECAI}, 2020.

\bibitem{guo2019single}
Zichao Guo, Xiangyu Zhang, Haoyuan Mu, Wen Heng, Zechun Liu, Yichen Wei, and
  Jian Sun.
\newblock Single path one-shot neural architecture search with uniform
  sampling.
\newblock {\em arXiv preprint arXiv:1904.00420}, 2019.

\bibitem{kaimingInit}
Kaiming He, X. Zhang, Shaoqing Ren, and Jian Sun.
\newblock Delving deep into rectifiers: Surpassing human-level performance on
  imagenet classification.
\newblock {\em 2015 IEEE International Conference on Computer Vision (ICCV)},
  2015.

\bibitem{he2016deep}
Kaiming He, Xiangyu Zhang, Shaoqing Ren, and Jian Sun.
\newblock Deep residual learning for image recognition.
\newblock In {\em CVPR}, 2016.

\bibitem{resnet}
Kaiming He, Xiangyu Zhang, Shaoqing Ren, and Jian Sun.
\newblock Deep residual learning for image recognition.
\newblock In {\em CVPR}, 2016.

\bibitem{AB}
Byeongho Heo, Minsik Lee, Sangdoo Yun, and Jin~Young Choi.
\newblock Knowledge transfer via distillation of activation boundaries formed
  by hidden neurons.
\newblock In {\em AAAI}, 2019.

\bibitem{ref10_kd}
Geoffrey Hinton, Oriol Vinyals, and Jeff Dean.
\newblock Distilling the knowledge in a neural network.
\newblock {\em arXiv preprint arXiv:1503.02531}, 2015.

\bibitem{kd}
Geoffrey Hinton, Oriol Vinyals, and Jeff Dean.
\newblock Distilling the knowledge in a neural network.
\newblock In {\em arXiv:1503.02531}, 2015.

\bibitem{li2021nas}
Yiming Hu, Xingang Wang, Lujun Li, and Qingyi Gu.
\newblock Improving one-shot nas with shrinking-and-expanding supernet.
\newblock {\em Pattern Recognition}, 2021.

\bibitem{ref40_nst}
Zehao Huang and Naiyan Wang.
\newblock Like what you like: Knowledge distill via neuron selectivity
  transfer.
\newblock {\em arXiv preprint arXiv:1707.01219}, 2017.

\bibitem{nst}
Zehao Huang and Naiyan Wang.
\newblock Like what you like: {K}nowledge distill via neuron selectivity
  transfer.
\newblock {\em arXiv:1707.01219}, 2017.

\bibitem{ref28_factor}
Jangho Kim, SeoungUk Park, and Nojun Kwak.
\newblock Paraphrasing complex network: Network compression via factor
  transfer.
\newblock In {\em NeurIPS}, 2018.

\bibitem{ref17_one}
Xu Lan, Xiatian Zhu, and Shaogang Gong.
\newblock Knowledge distillation by on-the-fly native ensemble.
\newblock In {\em NeurIPS}, 2018.

\bibitem{snip}
Namhoon Lee, Thalaiyasingam Ajanthan, and Philip Torr.
\newblock Snip: Single-shot network pruning based on connection sensitivity.
\newblock In {\em ICLR}, 2018.

\bibitem{Li2020BlockWiselySN}
Changlin Li, Jiefeng Peng, Liuchun Yuan, Guangrun Wang, Xiaodan Liang, Liang
  Lin, and Xiaojun Chang.
\newblock Block-wisely supervised neural architecture search with knowledge
  distillation.
\newblock {\em CVPR}, 2020.

\bibitem{li2022self}
Lujun Li.
\newblock Self-regulated feature learning via teacher-free feature
  distillation.
\newblock In {\em ECCV}, 2022.

\bibitem{lishadow}
Lujun Li and Zhe Jin.
\newblock Shadow knowledge distillation: Bridging offline and online knowledge
  transfer.
\newblock In {\em NeuIPS}, 2022.

\bibitem{li2022SFF}
Lujun Li, Liang Shiuan-Ni, Ya Yang, and Zhe Jin.
\newblock Boosting online feature transfer via separable feature fusion.
\newblock In {\em IJCNN}, 2022.

\bibitem{li2022tf}
Lujun Li, Liang Shiuan-Ni, Ya Yang, and Zhe Jin.
\newblock Teacher-free distillation via regularizing intermediate
  representation.
\newblock In {\em IJCNN}, 2022.

\bibitem{Li2019RandomSA}
Liam Li and Ameet~S. Talwalkar.
\newblock Random search and reproducibility for neural architecture search.
\newblock {\em ArXiv}, 2019.

\bibitem{li2020explicit}
Lujun Li, Yikai Wang, Anbang Yao, Yi Qian, Xiao Zhou, and Ke He.
\newblock Explicit connection distillation.
\newblock In {\em ICLR}, 2020.

\bibitem{ZenNAS}
Ming Lin, Pichao Wang, Zhenhong Sun, Hesen Chen, Xiuyu Sun, Qi Qian, Hao Li,
  and Rong Jin.
\newblock Zen-nas: A zero-shot nas for high-performance image recognition.
\newblock 2021.

\bibitem{lin2022knowledge}
Sihao Lin, Hongwei Xie, Bing Wang, Kaicheng Yu, Xiaojun Chang, Xiaodan Liang,
  and Gang Wang.
\newblock Knowledge distillation via the target-aware transformer.
\newblock In {\em CVPR}, 2022.

\bibitem{darts}
Hanxiao Liu, Karen Simonyan, and Yiming Yang.
\newblock {DARTS:} differentiable architecture search.
\newblock In {\em ICLR}.

\bibitem{liu2018darts}
Hanxiao Liu, Karen Simonyan, and Yiming Yang.
\newblock Darts: Differentiable architecture search.
\newblock {\em arXiv preprint arXiv:1806.09055}, 2018.

\bibitem{ickd}
Li Liu, Qinwen Huang, Sihao Lin, Hongwei Xie, Bing Wang, Xiaojun Chang, and
  Xiao-Xue Liang.
\newblock Exploring inter-channel correlation for diversity-preserved knowledge
  distillation.
\newblock {\em 2021 ICCV}, 2021.

\bibitem{ref22_search_student}
Yu Liu, Xuhui Jia, Mingxing Tan, Raviteja Vemulapalli, Yukun Zhu, Bradley
  Green, and Xiaogang Wang.
\newblock Search to distill: Pearls are everywhere but not the eyes.
\newblock In {\em CVPR}, 2020.

\bibitem{lopez2015unifying}
David Lopez-Paz, L{\'e}on Bottou, Bernhard Sch{\"o}lkopf, and Vladimir Vapnik.
\newblock Unifying distillation and privileged information.
\newblock {\em arXiv preprint arXiv:1511.03643}, 2015.

\bibitem{NWOT}
Joe Mellor, Jack Turner, Amos Storkey, and Elliot~J Crowley.
\newblock Neural architecture search without training.
\newblock In {\em ICML}, 2021.

\bibitem{ATKD}
Seyed~Iman Mirzadeh, Mehrdad Farajtabar, Ang Li, Nir Levine, Akihiro Matsukawa,
  and Hassan Ghasemzadeh.
\newblock Improved knowledge distillation via teacher assistant.
\newblock In {\em AAAI}, 2020.

\bibitem{rkd}
Wonpyo Park, Dongju Kim, Yan Lu, and Minsu Cho.
\newblock Relational knowledge distillation.
\newblock In {\em CVPR}, 2019.

\bibitem{ref13_rkd}
Wonpyo Park, Yan Lu, Minsu Cho, and Dongju Kim.
\newblock Relational knowledge distillation.
\newblock In {\em CVPR}, 2019.

\bibitem{ref42_pkt}
Nikolaos Passalis and Anastasios Tefas.
\newblock Learning deep representations with probabilistic knowledge transfer.
\newblock In {\em ECCV}, 2018.

\bibitem{pkt}
Nikolaos Passalis and Anastasios Tefas.
\newblock Learning deep representations with probabilistic knowledge transfer.
\newblock In {\em ECCV}, 2018.

\bibitem{cc}
Baoyun Peng, Xiao Jin, Jiaheng Liu, Dongsheng Li, Yichao Wu, Yu Liu, Shunfeng
  Zhou, and Zhaoning Zhang.
\newblock Correlation congruence for knowledge distillation.
\newblock In {\em ICCV}, 2019.

\bibitem{ref40_cc}
Baoyun Peng, Xiao Jin, Jiaheng Liu, Shunfeng Zhou, Yichao Wu, Yu Liu,
  Dong-sheng Li, and Zhaoning Zhang.
\newblock Correlation congruence for knowledge distillation.
\newblock In {\em ICCV}, 2019.

\bibitem{peng2020cream}
Houwen Peng, Hao Du, Hongyuan Yu, Qi Li, Jing Liao, and Jianlong Fu.
\newblock Cream of the crop: Distilling prioritized paths for one-shot neural
  architecture search.
\newblock {\em NeurIPS}, 2020.

\bibitem{ENAS}
Hieu Pham, Melody~Y. Guan, Barret Zoph, Quoc~V. Le, and Jeff Dean.
\newblock Efficient neural architecture search via parameter sharing.
\newblock In {\em ICML}, 2018.

\bibitem{li2021seg}
Jie Qin, Jie Wu, Xuefeng Xiao, Lujun Li, and Xingang Wang.
\newblock Activation modulation and recalibration scheme for weakly supervised
  semantic segmentation.
\newblock In {\em AAAI}, 2022.

\bibitem{Radosavovic2019OnND}
Ilija Radosavovic, Justin Johnson, Saining Xie, Wan-Yen Lo, and Piotr
  Doll{\'a}r.
\newblock On network design spaces for visual recognition.
\newblock {\em 2019 ICCV}, 2019.

\bibitem{Radosavovic2020DesigningND}
Ilija Radosavovic, Raj~Prateek Kosaraju, Ross~B. Girshick, Kaiming He, and
  Piotr Doll{\'a}r.
\newblock Designing network design spaces.
\newblock {\em CVPR}, 2020.

\bibitem{ref11_feature_kd}
Adriana Romero, Nicolas Ballas, Samira~Ebrahimi Kahou, Antoine Chassang, Carlo
  Gatta, and Yoshua Bengio.
\newblock Fitnets: Hints for thin deep nets.
\newblock In {\em ICLR}, 2015.

\bibitem{fitnets}
Adriana Romero, Nicolas Ballas, Samira~Ebrahimi Kahou, Antoine Chassang, Carlo
  Gatta, and Yoshua Bengio.
\newblock Fitnets: {H}ints for thin deep nets.
\newblock {\em ICLR}, 2015.

\bibitem{Grad}
R.~R. Selvaraju, Abhishek Das, Ramakrishna Vedantam, Michael Cogswell, D.
  Parikh, and Dhruv Batra.
\newblock Grad-cam: Visual explanations from deep networks via gradient-based
  localization.
\newblock 2019.

\bibitem{jacob}
Suraj Srinivas and Fran{\c{c}}ois Fleuret.
\newblock Knowledge transfer with jacobian matching.
\newblock In {\em ICML}, 2018.

\bibitem{syflow}
Hidenori Tanaka, Daniel Kunin, Daniel~L Yamins, and Surya Ganguli.
\newblock Pruning neural networks without any data by iteratively conserving
  synaptic flow.
\newblock {\em NeurIPS}, 2020.

\bibitem{ref14_relation_crd}
Yonglong Tian, Dilip Krishnan, and Phillip Isola.
\newblock Contrastive representation distillation.
\newblock In {\em ICLR}, 2020.

\bibitem{crd}
Yonglong Tian, Dilip Krishnan, and Phillip Isola.
\newblock Contrastive representation distillation.
\newblock In {\em ICLR}, 2020.

\bibitem{ref41_sp}
Frederick Tung and Greg Mori.
\newblock Similarity-preserving knowledge distillation.
\newblock In {\em ICCV}, 2019.

\bibitem{sp}
Frederick Tung and Greg Mori.
\newblock Similarity-preserving knowledge distillation.
\newblock In {\em ICCV}, 2019.

\bibitem{vapnik1998statistical}
Vladimir Vapnik.
\newblock {\em Statistical learning theory. 1998}.

\bibitem{wangdionysus}
Likang Wang and Lei Chen.
\newblock Dionysus: Recovering scene structures by dividing into semantic
  pieces.

\bibitem{wang2023ftso}
Likang Wang and Lei Chen.
\newblock Ftso: Effective nas via first topology second operator.
\newblock 2023.

\bibitem{wang2022mvsnet}
Likang Wang, Yue Gong, Xinjun Ma, Qirui Wang, Kaixuan Zhou, and Lei Chen.
\newblock Is-mvsnet: Importance sampling-based mvsnet.
\newblock In {\em ECCV}, 2022.

\bibitem{wang2023flora}
Likang Wang, Yue Gong, Qirui Wang, Kaixuan Zhou, and Lei Chen.
\newblock Flora: dual-frequency loss-compensated real-time monocular 3d video
  reconstruction.
\newblock In {\em AAAI}, 2023.

\bibitem{scda}
Xiu-Shen Wei, Jian-Hao Luo, Jianxin Wu, and Zhi-Hua Zhou.
\newblock Selective convolutional descriptor aggregation for fine-grained image
  retrieval.
\newblock {\em TIP}, 2017.

\bibitem{wei2022convformer}
Zimian Wei, Hengyue Pan, Lujun~Li Li, Menglong Lu, Xin Niu, Peijie Dong, and
  Dongsheng Li.
\newblock Convformer: Closing the gap between cnn and vision transformers.
\newblock {\em arXiv preprint arXiv:2209.07738}, 2022.

\bibitem{wu2019fbnet}
Bichen Wu, Xiaoliang Dai, Peizhao Zhang, Yanghan Wang, Fei Sun, Yiming Wu,
  Yuandong Tian, Peter Vajda, Yangqing Jia, and Kurt Keutzer.
\newblock Fbnet: Hardware-aware efficient convnet design via differentiable
  neural architecture search.
\newblock In {\em CVPR}, 2019.

\bibitem{li2022norm}
Liu Xiaolong, Li Lujun, Li Chao, and Anbang Yao.
\newblock Norm: Knowledge distillation via n-to-one representation matching.
\newblock In {\em ICLR}, 2023.

\bibitem{knas}
Jingjing Xu, Liang Zhao, Junyang Lin, Rundong Gao, Xu Sun, and Hongxia Yang.
\newblock Knas: Green neural architecture search.
\newblock In {\em ICML}, 2021.

\bibitem{yang2021knowledge}
Jing Yang, Brais Martinez, Adrian Bulat, Georgios Tzimiropoulos, et~al.
\newblock Knowledge distillation via softmax regression representation
  learning.
\newblock ICLR, 2021.

\bibitem{ref12_attention_kd}
Sergey Zagoruyko and Nikos Komodakis.
\newblock Paying more attention to attention: Improving the performance of
  convolutional neural networks via attention transfer.
\newblock In {\em ICLR}, 2017.

\bibitem{at}
Sergey Zagoruyko and Nikos Komodakis.
\newblock Paying more attention to attention: {I}mproving the performance of
  convolutional neural networks via attention transfer.
\newblock {\em ICLR}, 2017.

\bibitem{ref15_dml}
Ying Zhang, Tao Xiang, Timothy~M Hospedales, and Huchuan Lu.
\newblock Deep mutual learning.
\newblock In {\em CVPR}, 2018.

\bibitem{zhao2022decoupled}
Borui Zhao, Quan Cui, Renjie Song, Yiyu Qiu, and Jiajun Liang.
\newblock Decoupled knowledge distillation.
\newblock In {\em CVPR}, 2022.

\bibitem{Zheng2022NeuralAS}
Xiawu Zheng, Xiang Fei, Lei Zhang, Chenglin Wu, Fei Chao, Jianzhuang Liu, Wei
  Zeng, Yonghong Tian, and Rongrong Ji.
\newblock Neural architecture search with representation mutual information.
\newblock {\em CVPR}, 2022.

\bibitem{zhou2016learning}
Bolei Zhou, Aditya Khosla, Agata Lapedriza, Aude Oliva, and Antonio Torralba.
\newblock Learning deep features for discriminative localization.
\newblock In {\em CVPR}, 2016.

\bibitem{nasnet}
Barret Zoph, Vijay Vasudevan, Jonathon Shlens, and Quoc~V Le.
\newblock Learning transferable architectures for scalable image recognition.
\newblock In {\em CVPR}, 2018.

\end{thebibliography}
}

\end{document}